\documentclass[10pt,twocolumn,letterpaper]{article}

\usepackage{iccv}      

\definecolor{iccvblue}{rgb}{0.21,0.49,0.74}
\usepackage[pagebackref,breaklinks,colorlinks,allcolors=iccvblue]{hyperref}

\newcommand\B[1]{{#1}}
\usepackage{newfloat}
\usepackage{listings}
\usepackage{adjustbox}
\usepackage{graphicx}
\usepackage{booktabs} 
\usepackage{subcaption} 
\usepackage{subcaption,booktabs}
\usepackage{color}
\usepackage[dvipsnames]{xcolor}
\usepackage{newfloat}
\usepackage{listings}
\usepackage{adjustbox}
\usepackage{multirow}

\def\paperID{*****} 
\def\confName{ICCV}
\def\confYear{2025}

\title{Constraining Multimodal Distribution for Domain Adaptation in Stereo Matching}

\author{Zhelun Shen$^{1,}$\thanks{Equal contribution.}
\;
Zhuo Li$^{2,*}$
\;
Chenming Wu$^{1,}$\thanks{Corresponding author, email: {\tt\small wuchenming@baidu.com}.}
\;
Zhibo Rao$^{3}$ 
\;
Lina Liu$^{4}$
\;
Yuchao Dai$^{5}$
\;
Liangjun Zhang$^{1}$
\vspace{0.2cm}
\\
{\normalsize $^{1}$ RAL, Baidu Research \quad \normalsize $^{2}$ ICT, Chinese Academy of Science} \\
{\normalsize \quad $^{3}$ Nanchang Hangkong University} 
{\normalsize \quad $^{4}$  Zhejiang University}
{\normalsize \quad $^{5}$ Northwestern Polytechnical University}
\vspace{-10pt}
}

\begin{document}
\maketitle

\begin{abstract}
    Recently, learning-based stereo matching methods have achieved great improvement in public benchmarks, where soft argmin and smooth L1 loss play a core contribution to their success. However, in unsupervised domain adaptation scenarios, we observe that these two operations often yield multimodal disparity probability distributions in target domains, resulting in degraded generalization. In this paper, we propose a novel approach, \underline{C}onstrain \underline{M}ulti-modal \underline{D}istribution (CMD), to address this issue. Specifically, we introduce \textit{uncertainty-regularized minimization} and \textit{anisotropic soft argmin} to encourage the network to produce predominantly unimodal disparity distributions in the target domain, thereby improving prediction accuracy. Experimentally, we apply the proposed method to multiple representative stereo-matching networks and conduct domain adaptation from synthetic data to unlabeled real-world scenes. Results consistently demonstrate improved generalization in both top-performing and domain-adaptable stereo-matching models. The code for CMD will be available at: \href{https://github.com/gallenszl/CMD}{https://github.com/gallenszl/CMD}.
\end{abstract}

\section{Introduction}

Stereo matching, the task of estimating a disparity or depth map from a pair of stereo images, is a foundational component in various computer vision applications, including autonomous driving~\cite{autonomousdriving, wang2023digging}, robot navigation~\cite{roboticsnavigation, salmeron2015tradeoff}, industrial automation~\cite{li2015stereo}, active 3D reconstruction~\cite{sui2020active}, and simultaneous localization and mapping (SLAM)~\cite{slam1, hussain2020s, slam2}. Recently, learning-based stereo matching methods have achieved state-of-the-art performance on public benchmarks~\cite{psmnet, gwcnet, ganet}. However, these methods face significant challenges when deployed in real-world environments, primarily due to the domain gap between training data (typically synthetic) and the target data (real-world scenes).

\B{The domain gap arises because collecting annotated real-world data is labor-intensive and costly, often requiring expensive LiDAR sensors and precise calibration. Consequently, synthetic datasets, which are easier to generate and annotate, are commonly used for training stereo matching models. However, a major limitation of this approach is that models trained on synthetic data often struggle to generalize well to real-world scenes~\cite{gcnet, psmnet}. This domain gap is particularly problematic because synthetic and real-world data often differ in texture, illumination, and noise characteristics, leading to a degradation in performance when models are tested on real-world data.}

\B{A key issue we observe in unsupervised domain adaptation for stereo matching is the tendency of models to produce multimodal disparity probability distributions in the target domain. These multimodal distributions arise when the network is unsure about the correct disparity, leading to multiple possible disparity modes rather than a single, sharp peak centered on the true disparity. This behavior significantly hinders generalization, as the network fails to produce accurate, unimodal disparity predictions in the target domain. Previous works have primarily relied on fine-tuning models with some labeled real-world data to address this issue~\cite{gcnet, psmnet}; however, this is not feasible when labeled real-world data is unavailable, as is often the case in practical applications.}

\B{In this paper, we address the domain adaptation challenge in stereo matching by focusing on two key operations in learning-based stereo matching: the \textit{soft argmin} operation~\cite{gcnet} and the \textit{smooth L1 loss}~\cite{psmnet}. While these components have been crucial for the success of supervised stereo matching, their impact on domain adaptation has not been fully explored. Specifically, these two operations only constrain the predicted disparity without directly influencing the underlying disparity probability distribution, making it possible for different disparity distributions (including multimodal ones) to result in the same predicted disparity. For better domain adaptation performance, it is critical to enforce the network to output a unimodal disparity probability distribution centered on the true disparity, thus improving the network’s confidence in its predictions.}

\begin{figure*}[!t]
\centering
\tabcolsep=0.05cm
\includegraphics[width=0.99\linewidth]{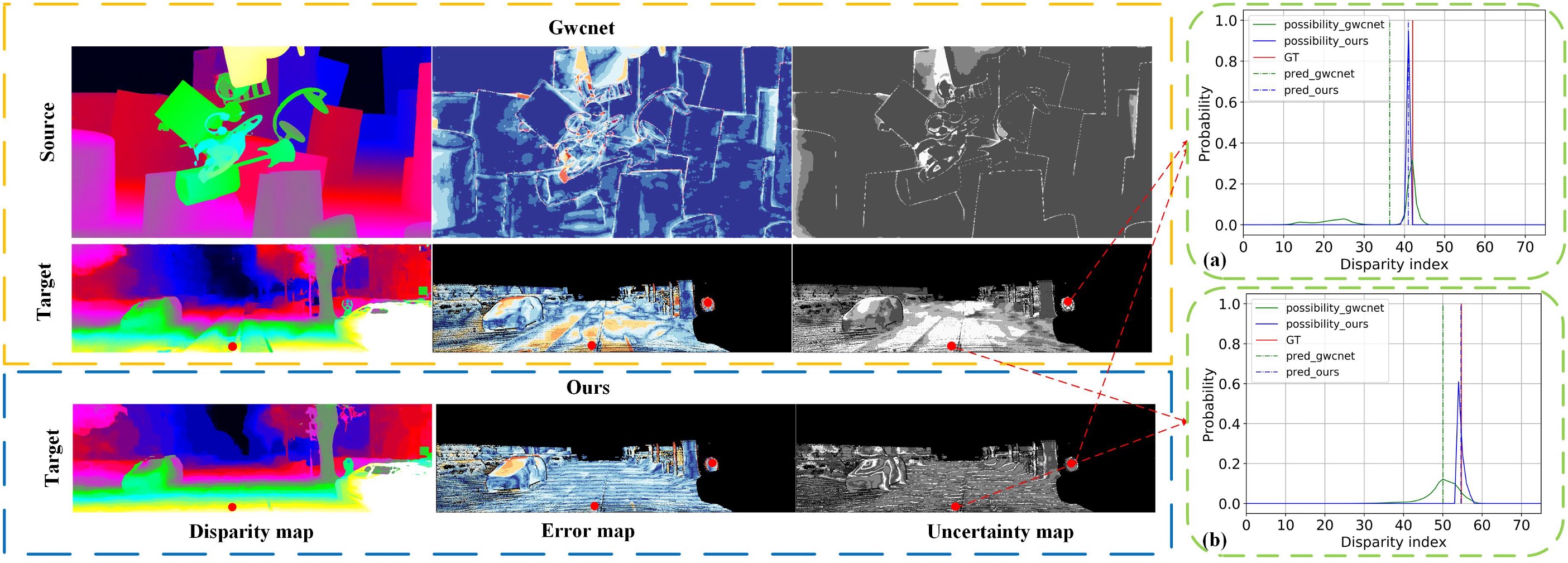}
\caption{\textbf{Left:} Proposed unsupervised domain adaptation for stereo matching (from left to right: disparity map, error map, and uncertainty map). Red and white denote large errors and high uncertainty. \textbf{Right:} Disparity probability distributions for two selected pixels (red point in the picture). Pred denotes the predicted disparity in this pixel and possibility denotes the corresponding disparity probability distribution.}
\label{fig: intro_visual}
\end{figure*}

\B{To tackle this issue, we propose two complementary techniques: \textit{uncertainty regularized minimization} and \textit{anisotropic soft argmin}, both designed to reduce the multimodal nature of disparity distributions in the target domain and improve performance on unlabelled target data.
First, we introduce \textit{uncertainty-regularized minimization}, a method inspired by the observation that uncertainty measures can serve as a proxy for the sharpness of the disparity probability distribution. The intuition here is that a sharp, unimodal distribution corresponds to low uncertainty, while a multimodal or flat distribution exhibits high uncertainty. By integrating an uncertainty-based loss term, we explicitly encourage the network to produce unimodal distributions. Importantly, this uncertainty loss does not require labeled ground-truth data, making it well-suited for unsupervised domain adaptation scenarios.}

Second, we propose \textit{anisotropic soft argmin}, a modification of the traditional soft argmin operation. The standard softmax function used in soft argmin converts the cost distribution into a probability distribution, but the sharpness of the resulting distribution is determined by the temperature parameter. We introduce an anisotropic version of soft argmin, where the temperature parameter can be tuned to sharpen the probability distribution, forcing the network to focus more on the correct disparity and penalizing incorrect predictions more strongly. This approach is particularly effective in driving faster convergence to the correct distribution and complements the uncertainty-regularized minimization strategy. Furthermore, anisotropic soft argmin also imposes a stronger penalty for false disparity estimations, facilitating faster convergence towards the correct distribution. Moreover, the proposed two operations also complement the pseudo-label-based method \cite{shen2023digging, unsuperviseddomainadaptation} and can further improve domain adaptation performance by self-distilling more accurate pseudo-labels.

\B{Experimentally, we apply the proposed method to multiple representative stereo matching networks \cite{gwcnet, itsa} and conduct domain adaptation from synthetic data to unlabeled real-world scenes. Corresponding results demonstrate that the proposed method can consistently improve the performance of top-performance and domain-generalized stereo matching methods on unlabeled target data. Moreover, multiple comparison experiments are conducted to verify that the proposed method can alleviate the multimodal distribution issue rather than just improve the performance by a black box. Fig.~\ref{fig: intro_visual} presents a visualization of the results, which showcases the consistent improvement in the sharpness of the distribution (measured by uncertainty) and the overall better performance achieved by our method. Its sub-figures (a) and (b) provide detailed examples illustrating how the proposed approach effectively constrains the multimodal distribution of the networks, ultimately leading to enhanced performance.}
In summary, the main contributions of this paper are as follows:

\begin{itemize}
    \item We identify and analyze the issue of multimodal disparity probability distributions in stereo matching when adapting to real-world domains, which leads to degraded generalization performance.
    \item We propose two simple yet effective solutions: \textit{uncertainty-regularized minimization} and \textit{anisotropic soft argmin}, both aimed at enforcing unimodal disparity distributions in the target domain.
    \item The proposed two operations are also complementary to the pseudo-label-based method and can improve the quality of generated pseudo-labels for further domain adaptation.
    \item Experimental results show that our approach can be applied to a wide range of stereo matching networks, consistently improving domain adaptation performance and alleviating the multimodal distribution issue.
\end{itemize}
\section{Related Work}

\subsection{Deep Stereo Matching}

The introduction of DispNet~\cite{dispnet} marked a significant milestone in stereo matching by pioneering the use of end-to-end deep learning models, which rely purely on data without the need for traditional stereo matching pipelines. While DispNet demonstrated the potential of data-driven approaches, training such models proved to be challenging due to the complexity of the stereo matching task. 

GCNet~\cite{gcnet} further advanced the field by introducing a differentiable 4D cost volume and the soft argmin operation, which incorporated traditional stereo matching principles into deep learning models. These modules not only improved accuracy but also simplified the training process by leveraging prior knowledge from classical stereo matching algorithms. Building on this foundation, subsequent works have proposed a variety of novel architectures and techniques to push the state of the art even further.

For example, Kim \textit{et al.}~\cite{kim2018unified} introduced a semi-supervised architecture that estimates stereo confidence by using both the matching cost volume and the disparity map as inputs. Zeng \textit{et al.}~\cite{zeng2021deep} proposed hysteresis attention and supervised cost volume construction to enhance stereo matching performance. Zhang \textit{et al.}~\cite{zhang2023active} tackled the limitations of random disparity sampling by introducing active disparity sampling with an adjoint network, while Chen \textit{et al.}~\cite{chen2024mocha} introduced a motif channel correlation volume for more accurate edge-matching costs. Li \textit{et al.}~\cite{li2024local} leveraged local structure information (LSI) to handle challenging regions such as object boundaries, and Gong \textit{et al.}~\cite{gong2024learning} employed interest points to impose geometric constraints on the stereo matching process.

The various advancements in stereo matching can be broadly classified into five key categories: 
1) \textbf{Feature representation}~\cite{emcua, mcvmfc, cheng2021two, gong2024learning, zeng2021deep}, 
2) \textbf{Cost volume enhancement}~\cite{gwcnet, cascade, parameterized, chen2024mocha, zeng2021deep}, 
3) \textbf{Cost aggregation strategies}~\cite{ganet, psmnet, lipson2021raft, li2022practical, xu2023iterative}, 
4) \textbf{Disparity computation techniques}~\cite{oversmooth, pds}, and 
5) \textbf{Multi-task learning frameworks}~\cite{masked, elfnet, chen2023learning, kim2018unified}. 
Each of these contributions has pushed the boundaries of performance and operational efficiency in stereo matching.

More closely related to our work is ACFNet~\cite{acfnet}, which directly constrains the disparity probability distribution by enforcing unimodality through cross-entropy loss. Follow-up work, such as~\cite{xu2024adaptive}, proposed an adaptive multimodal cross-entropy loss to better model the distribution. However, these methods primarily focus on the source domain and overlook the multimodal disparity distributions that arise in the target domain during domain adaptation. In contrast, we propose \textit{uncertainty-regularized minimization} and \textit{anisotropic soft argmin} to directly constrain the disparity probability distribution in the target domain, significantly improving generalization and enhancing the quality of pseudo-labels for domain adaptation.

\subsection{Domain Adaptation}

\B{Deep domain adaptation has emerged as a powerful learning technique to address the challenge of limited labeled data. Four major approaches to reduce domain shifts are class criterion \cite{class1, class2}, statistic criterion \cite{statistic1}, architecture criterion \cite{network1}, and geometric criterion \cite{geometric}. Recently, several works have provided new insights into this task. Specifically, \cite{LRCA} proposes a low-rank correlation learning (LRCL) method to alleviate the disturbances and negative effects of noise introduced by the used domain data. \cite{DIA} introduces a discriminative invariant alignment (DIA) approach to better align domain distributions across different datasets. \cite{GDCSL} presents a guided discrimination and correlation subspace learning (GDCSL) method to learn domain-invariant, category-discriminative, and highly correlated features. \cite{IDKC} proposes an incremental discriminative knowledge consistency (IDKC) method that integrates cross-domain mapping, distribution matching, discriminative knowledge preservation, and domain-specific geometry structure consistency into a unified learning model for improved heterogeneous domain adaptation. \cite{cdsl} introduces a cross-domain structure learning (CDSL) method that combines global distribution alignment and local discriminative structure preservation to better extract domain-invariant features. However, most previous works have focused on image classification tasks. In this paper, we primarily focus on domain adaptation in stereo matching.}

\subsection{Robust Stereo Matching}

In the context of robust stereo matching—specifically for domain generalization and adaptation—prior work can be broadly divided into three main approaches, each targeting different levels of alignment between the source and target domains in the stereo matching pipeline.

The first approach focuses on aligning the source and target domains at the input image level. This is typically achieved through methods such as generative adversarial network (GAN)-based image-to-image translation~\cite{stereogan} or non-adversarial progressive color transfer algorithms~\cite{adastereo_ijcv}, both of which aim to adapt the appearance of source domain images to resemble those of the target domain.

The second approach aims to align the source and target domains at the internal feature level, where invariant and informative features are extracted across domains. Techniques such as information-theoretic shortcut avoidance~\cite{itsa}, broad-spectrum representation learning~\cite{graftnet}, feature consistency learning~\cite{FeatureConsistency}, and trainable non-local graph-based filters~\cite{dsmnet} have been proposed to achieve this goal.

The third approach focuses on aligning the source and target domains at the output space, often through self-supervised learning~\cite{adastereo_ijcv, madnet} or pseudo-label knowledge distillation~\cite{unsuperviseddomainadaptation}. These techniques leverage unlabeled target domain data to generate reliable pseudo-labels, which can then be used to refine the model and improve domain adaptation performance.

Despite the progress made by these approaches, they often overlook intrinsic challenges specific to stereo matching. For instance, two core operations in supervised stereo matching—soft argmin and smooth L1 loss—tend to produce multimodal disparity distributions \B{in the target domain}, leading to degraded generalization. In this work, we directly address this problem by proposing techniques to constrain multimodal disparity distributions in the target domain, thereby improving generalization and producing higher-quality pseudo-labels for further domain adaptation.
\section{Our Approach}

\subsection{Preliminaries}

Current end-to-end deep stereo matching methods typically follow a four-stage pipeline: feature extraction, cost volume construction, cost aggregation, and disparity computation~\cite{survey}. Let $f_\theta$ denote the feature extraction network, $\zeta$ the cost volume construction, and $\delta$ the cost aggregation network. A standard stereo matching network can then be formulated as:
\begin{eqnarray}
d = \text{soft} \arg \min (\delta (\zeta (f_\theta(I_l), f_\theta(I_r)))),
\end{eqnarray}
where $I_l$ and $I_r$ are the left and right stereo images, respectively. In most modern deep stereo matching networks, the \textit{soft} argmin operation is used to compute the final disparity. Our method is compatible with any network that adopts \textit{soft} argmin for disparity estimation.

In Section~\ref{Experiment}, we will demonstrate how our approach consistently enhances generalization across various state-of-the-art stereo matching methods, particularly in domain-adaptation scenarios.

\subsection{Multimodal Distribution Analysis and Metric}

The \textit{soft} argmin operation~\cite{gcnet} is widely employed in stereo matching to compute disparity from a probability distribution. It can be written as:
\begin{eqnarray}
\hat{d} = \sum_{i=0}^{D_{\max}} i \times p(i),
\end{eqnarray}
where $p(i) = \sigma(-C(i))$, $\sigma$ is the softmax function, $C(i)$ is the cost distribution, and $p(i)$ represents the probability at candidate disparity index $i$. 

To supervise the predicted disparity, the smooth L1 loss is typically used:
\begin{eqnarray}
L_s = \frac{1}{N} \sum_{n=1}^N \text{smooth}_{L_1}(d_n - \hat{d}_n),
\end{eqnarray}
where the smooth L1 loss is defined as:
\begin{eqnarray}
\text{smooth}_{L_1}(x) = \left\{
\begin{array}{ll}
0.5x^2, & \text{if } |x| < 1, \\
|x| - 0.5, & \text{otherwise}.
\end{array}
\right.
\end{eqnarray}
Here, $N$ denotes the valid pixels, and \B{$d_n$ and $\hat{d}_n$ are the valid area of ground truth and predicted disparities, respectively.}


\begin{table}[t!]
\centering
{
\caption{Distribution sharpness evaluation by uncertainty metric. SceneFlow and KITTI are used as source domain (SD) and target domain (TD), respectively. 
All metrics can obtain consistently higher uncertainty in the target domain, \textit{i.e.}, a higher degree toward the multimodal distribution. Instead, our method can greatly alleviate such issues.}
\label{tab:sharpness}
\resizebox{0.45\textwidth}{!}{ 
\begin{tabular}{c|c|cc|c}
\toprule
\multirow{2}{*}{Baseline} & Uncertainty & \multicolumn{2}{c|}{Original Implementation} & + Our Method \\ \cline{3-4} 
& Metric & \multicolumn{1}{c|}{SD} & TD & TD \\ \midrule
\multirow{3}{*}{Gwcnet}   & MSM & \multicolumn{1}{c|}{0.54} & 0.72 & 0.32 \\  
                          & Entropy & \multicolumn{1}{c|}{1.32} & 1.96 & 0.46 \\  
                          & PER & \multicolumn{1}{c|}{0.80} & 0.92 & 0.64 \\ 
\bottomrule
\end{tabular}
} 
}
\end{table}
However, the smooth L1 loss only supervises the predicted disparity $\hat{d}$, not the disparity probability distribution $p(i)$. This leads to an ill-posed problem: multiple disparity probability distributions can generate the same $\hat{d}$. Ideally, the disparity probability distribution should be unimodal and peaked at the true disparity. A multimodal distribution, on the other hand, significantly degrades performance. Prior work~\cite{gcnet} addresses this issue by relying on network regularization to produce unimodal distributions, but this assumption holds primarily when large amounts of labeled data are available (i.e., in the source domain). In unseen target domains, networks tend to produce multimodal disparity distributions, which hinder generalization.

To verify our hypothesis, we propose a quantitative evaluation to measure the tendency of the disparity probability distribution towards multimodality. Our metric, inspired by traditional stereo matching uncertainty measures~\cite{confidencesurvey}, evaluates the sharpness of the cost volume distribution. Sharper distributions indicate higher confidence. Since the cost distribution can be transformed into a probability distribution via the softmax, we repurpose uncertainty metrics to evaluate multimodality in the probability distribution. Specifically, we use the Matching Score Measure (MSM), Perturbation Measure (PER), and Entropy:
\begin{eqnarray}
\text{MSM}(p) = 1 - p(i_1),
\end{eqnarray}
\begin{eqnarray}
\text{Entropy}(p) = -\sum_{i=0}^{D_{\max}} p(i) \log p(i),
\end{eqnarray}
\begin{eqnarray}
\text{PER}(p) = \frac{1}{M} \sum_{i \neq i_1} \exp\left(-\frac{(p(i_1) - p(i))^2}{s^2}\right),
\end{eqnarray}
where $p(i)$ represents the probability at disparity index $i$, and $i_1$ denotes the index of highest probability. $M$ is the number of valid disparity indices.

\begin{figure*}[!t]
	\centering
	\tabcolsep=0.05cm
	\begin{tabular}{c c c}

	\includegraphics[width=0.33\linewidth]{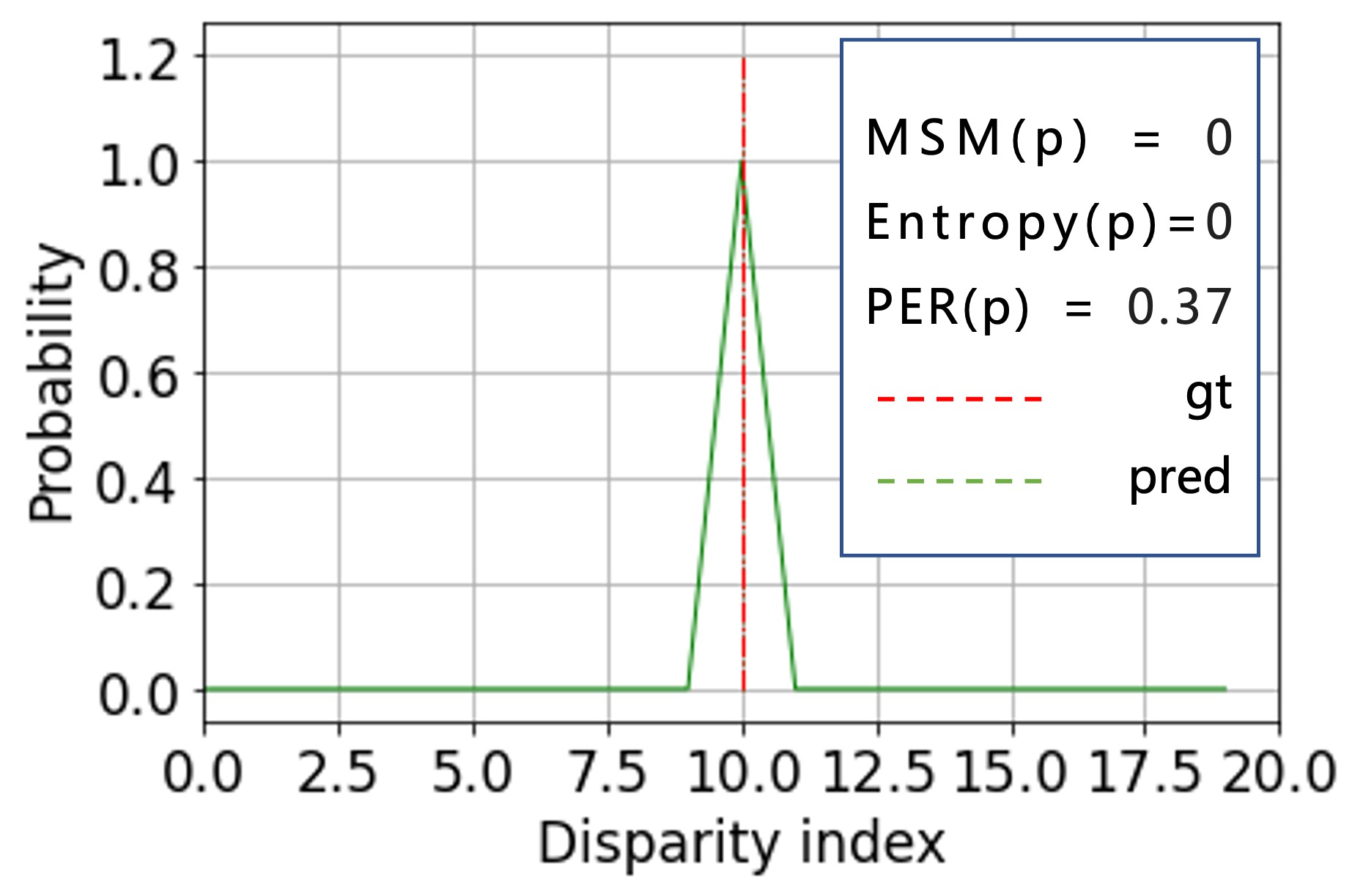}&
    \includegraphics[width=0.33\linewidth]{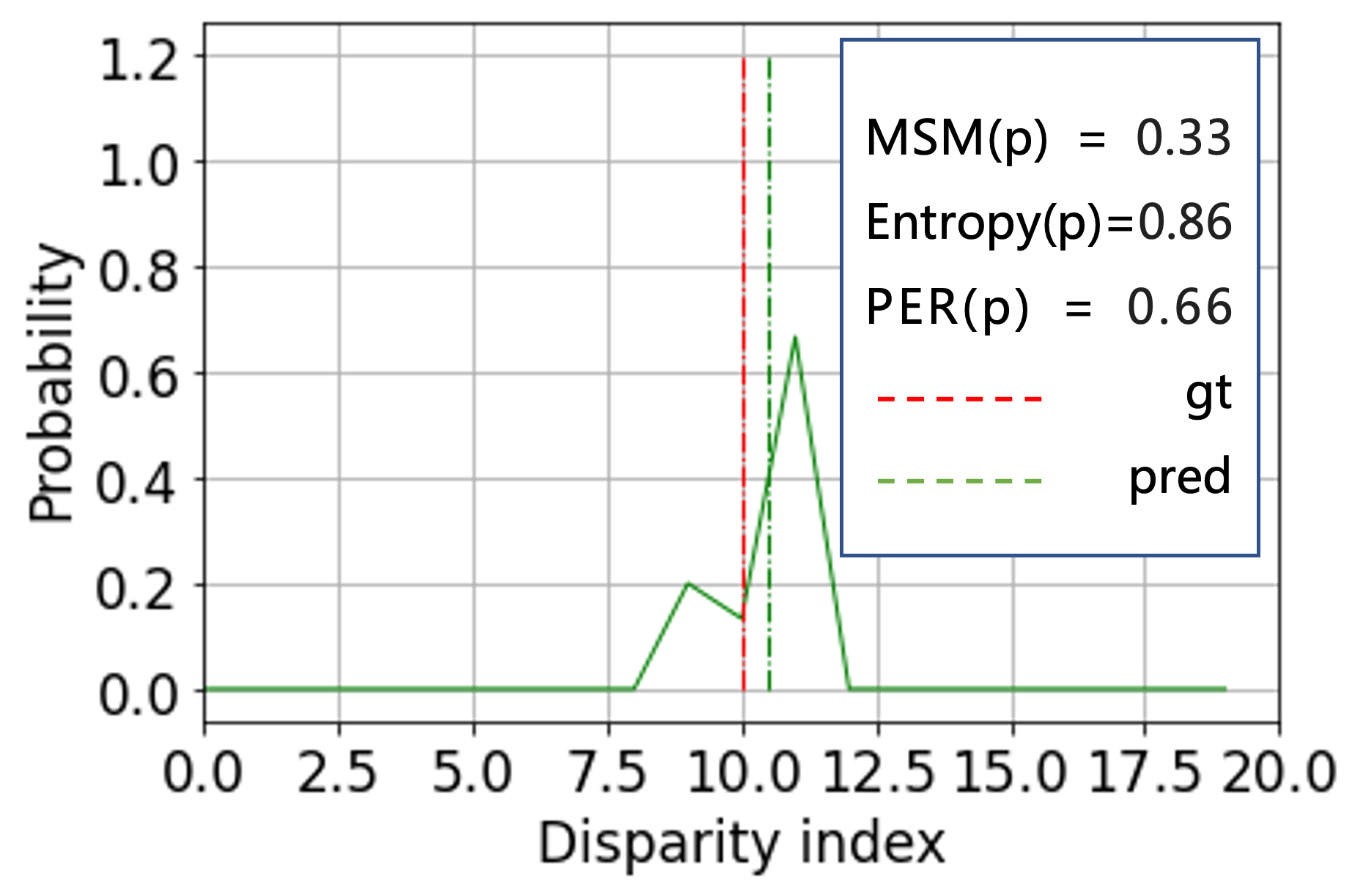}&
     \includegraphics[width=0.33\linewidth]{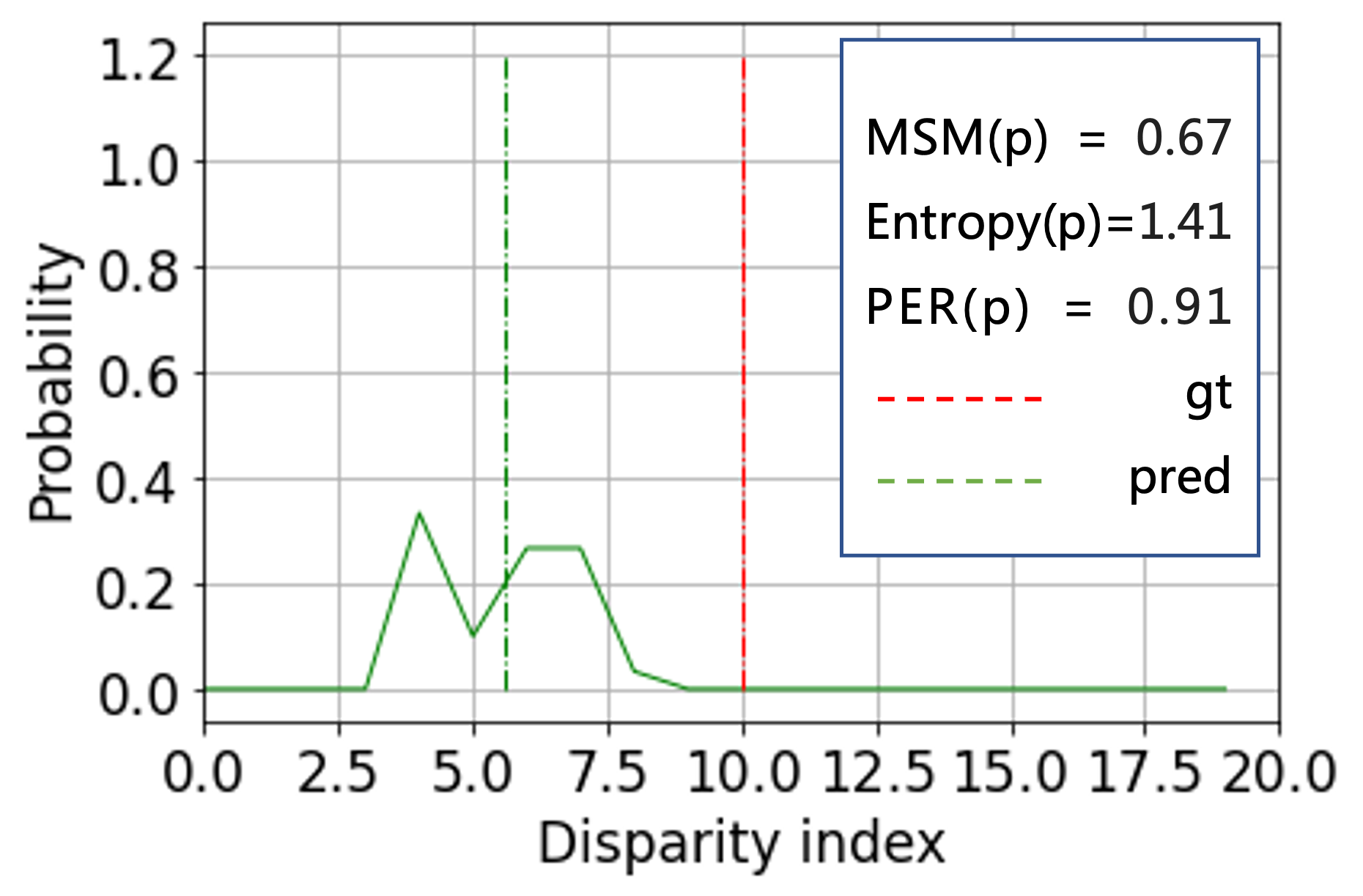}\\

    {\scriptsize(a) Unimodal} &  {\scriptsize(b) Predominantly unimodal} & {\scriptsize(c) Multimodal}

	\end{tabular}
	\caption{Some toy samples of distribution sharpness evaluation. Ground truth disparity is 10px. $pred$ is the predicted disparity. The disparity searching range is from 1 to 20 with 20 hypothesis planes.}
	\label{fig: uncertainty metric}
\end{figure*}

Among the evaluated uncertainty metrics, the Matching Score Measure (MSM) represents a local uncertainty measure, relying solely on the highest probability value within the cost distribution. In contrast, Perturbation Measure (PER) and entropy are global uncertainty measures that take into account the entire cost distribution to estimate uncertainty. Figure~\ref{fig: uncertainty metric} provides illustrative examples demonstrating the evaluation of distribution sharpness using these metrics. As depicted, the more the probability distribution shifts toward a multimodal form, the higher the associated uncertainty, which in turn leads to greater disparity estimation errors. This trend is observed consistently across all uncertainty metrics, which verifies the effectiveness of the proposed uncertainty metric.

The final multimodal distribution degree evaluation is shown in Tab. \ref{tab:sharpness}. As shown, for the original implementation, both local and global uncertainty measures exhibit consistently higher uncertainty in the target domain, indicating a stronger tendency toward multimodal distributions. For instance, the entropy in the target domain is 48.48\% higher than in the source domain. This significant increase further substantiates our hypothesis that existing methods often fail to adequately constrain the disparity probability distribution, resulting in a higher likelihood of producing multimodal distributions in unseen target domains and, consequently, poorer generalization performance. For a detailed evaluation of generalization, refer to Table~\ref{tab:constraint}.

\subsection{Uncertainty-Regularized Minimization}

To tackle the issue of multimodal distributions in the target domain, we propose a novel approach called \textit{Uncertainty-Regularized Minimization}. The key observation is that networks trained on the source domain produce low-uncertainty disparity maps in the source domain but exhibit high uncertainty in the target domain. Therefore, we introduce an additional loss term that penalizes high uncertainty, encouraging the network to output predominantly unimodal disparity distributions in the target domain.

Let $p^t$ represent the predicted disparity probability distribution in the target domain, and $U_m$ denote the chosen uncertainty metric (e.g., MSM, PER, Entropy). The uncertainty map $U(p^t)$ is computed as follows:
\begin{eqnarray}
U(p^t) = \left\{
\begin{array}{ll}
\text{MSM}(p^t), & \text{if } U_m == \text{MSM}, \\
\text{PER}(p^t), & \text{if } U_m == \text{PER}, \\
\text{Entropy}(p^t), & \text{if } U_m == \text{Entropy}.
\end{array}
\right.
\end{eqnarray}

We then calculate the mean of the uncertainty map to derive the uncertainty loss $L_u$:
\begin{eqnarray}
L_u(p^t) = \frac{1}{N} \sum U(p^t),
\end{eqnarray}
where $N$ represents the number of valid pixels. During training, the network is optimized by jointly minimizing the supervised smooth L1 loss $L_s$ on source images and the unsupervised uncertainty loss $L_u$ on target images. The overall loss function is:
\begin{eqnarray}
L = L_s(d^s, \hat{d}^s) + \lambda L_u(p^t),
\end{eqnarray}
where $\lambda$ is a balancing coefficient for the uncertainty loss, \B{$d^s$ is the ground truth disparity of the source domain,  $\hat{d}^s$ is the predicted disparity in the source domain.}


\B{\noindent\textbf{Convergence Analysis.}} 
The joint optimization framework achieves disparity concentration through two synergistic loss components:

Use Entropy Loss for example, as formalized in Eq.~\ref{L_u_ans}, we design an entropy minimization criterion:
\begin{equation}
L_u = -\sum_{i=0}^{D_{\max}} p(i) \log p(i)
\label{L_u_ans}
\end{equation}

This loss attains its minimum value 0 \textit{if} $\exists d \in \mathcal{D}, p(d) = 1$. Minimizing $L_u$ induces:
\begin{itemize}
    \item \textit{Mode Collapse}: Forces the probability mass to concentrate on a single disparity value
    \item \textit{Entropy Suppression}: Reduces distribution entropy from $\log(D_{\max}+1)$ (uniform) to 0 (deterministic)
    \item \textit{Unimodal Enforcement}: Eliminates secondary modes through gradient pressure $\frac{\partial L_u}{\partial p(i)} = -1 - \log p(i)$
\end{itemize}

Smooth Disparity Anchoring $L_s$: The smooth L1 loss anchors the concentration mode to the ground truth:
\begin{equation}
L_s = \frac{1}{n}\sum_{i=1}^n \begin{cases}
0.5(\hat{d}_i - d_i)^2, & |\hat{d}_i - d_i| < 1 \\
|\hat{d}_i - d_i| - 0.5, & \text{otherwise}
\end{cases}
\end{equation}

Convergence Dynamics: The compound loss $L = L_u + \lambda L_s$ ($\lambda$: trade-off parameter) generates dual-phase convergence:
\begin{enumerate}
    \item \textit{Concentration Phase}: $L_u$ dominates, reducing entropy $H(p^t)$ exponentially:
    \begin{equation}
    H(p^{(t)}) = H(p^{(0)})e^{-\gamma t}
    \end{equation}
    
    \item \textit{Alignment Phase}: $L_s$ dominates, guiding the concentration peak to $d$ with error decay:
    \begin{equation}
    \mathbb{E}[|\hat{d} - d|] = 0
    \end{equation}
\end{enumerate}

The distribution $p(d)$ evolves from uniform to unimodal, eventually converging to a Dirac delta distribution $\delta(d)$. This is formally equivalent to a normal distribution $\mathcal{N}(d, \sigma)$ with $\sigma \to 0$, giving the network both precise and confident disparity predictions.

\subsection{Anisotropic Soft Argmin}

To further control the sharpness of the disparity probability distribution, we propose a modified softmax function in the \textit{soft} argmin operation, which we call \textit{Anisotropic Soft Argmin}. The softmax function used in \textit{soft} argmin converts the cost distribution into a probability distribution as follows:
\begin{eqnarray}
p(i) = \frac{\exp(-C(i))}{\sum_{j=0}^{D_{\max}} \exp(-C(j))}.
\end{eqnarray}
To increase the sharpness of the disparity probability distribution, we propose scaling the exponential term by a temperature parameter $t > 1$, yielding the \textit{anisotropic softmax} function:
\begin{eqnarray}
p'(i) = \frac{\exp(-tC(i))}{\sum_{j=0}^{D_{\max}} \exp(-tC(j))}.
\end{eqnarray}
The corresponding disparity computation becomes:
\begin{eqnarray}
\hat{d}' = \sum_{i=0}^{D_{\max}} i \times p'(i).
\end{eqnarray}

\begin{figure*}[!t]
	\centering
	\tabcolsep=0.02cm
	\begin{tabular}{c c}
	\includegraphics[width=0.42\linewidth]{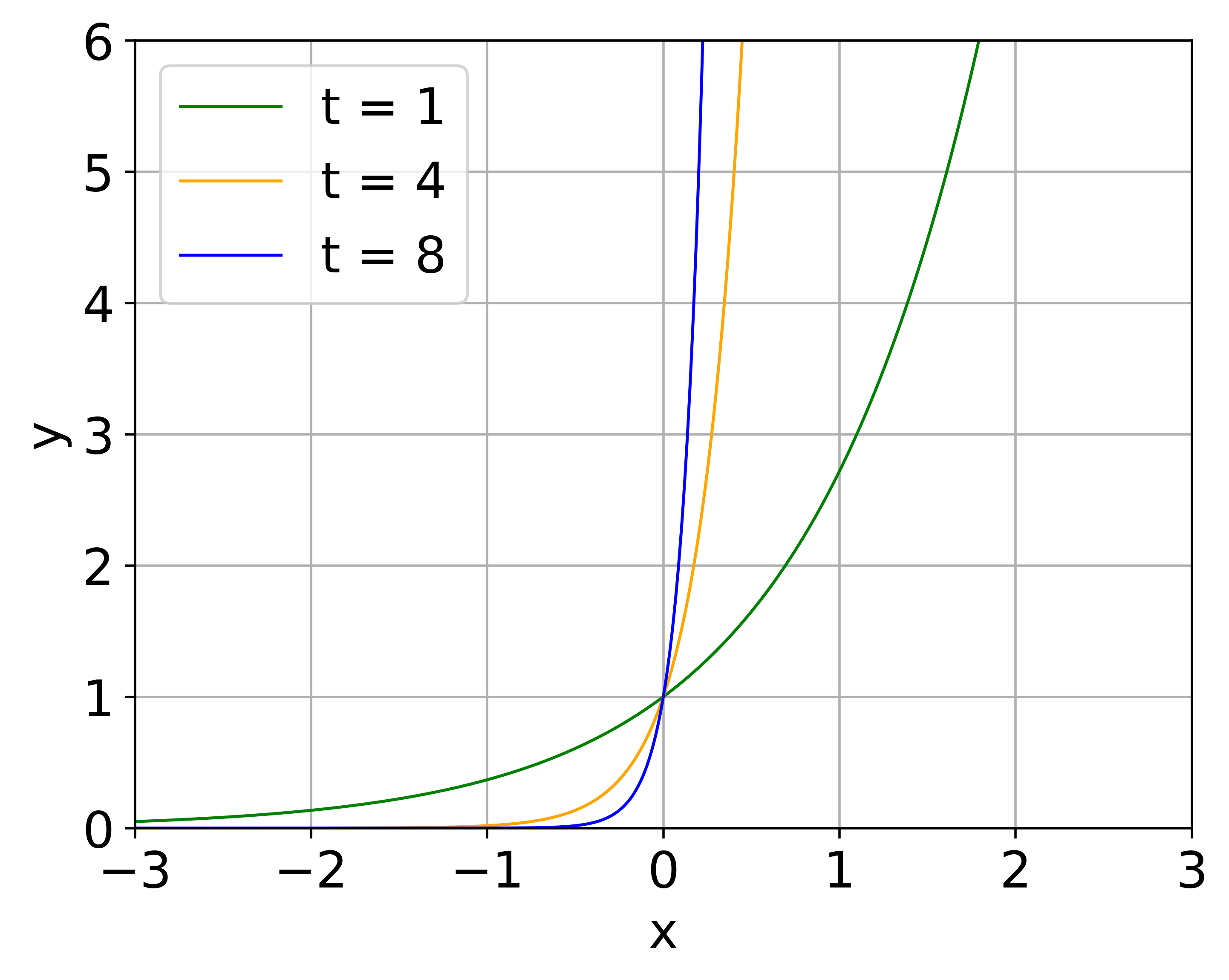}&
     \includegraphics[width=0.43\linewidth]{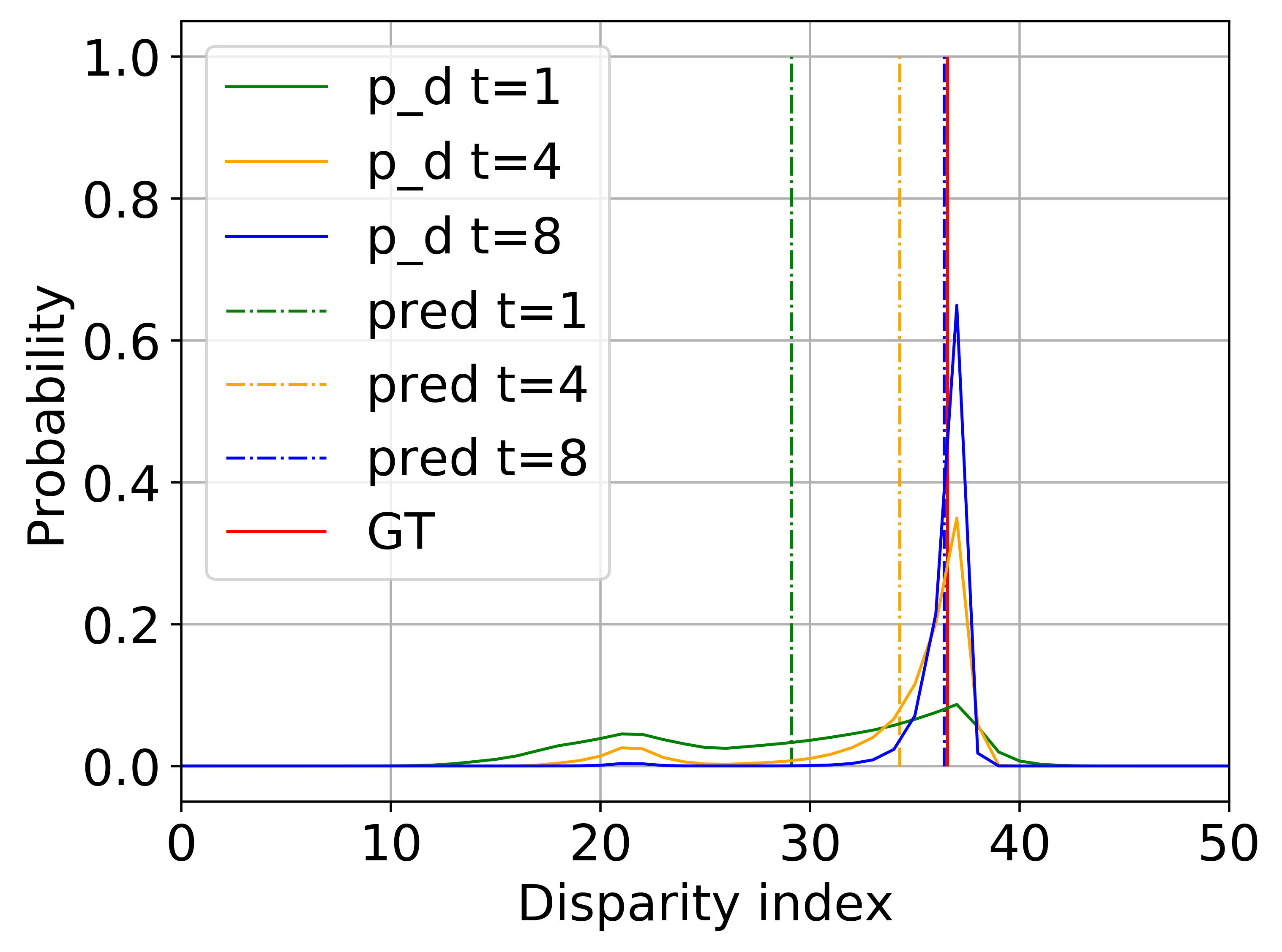}\\
	{(a) $f(x) = {e^{tx}}$} &	{(b)  Probability distribution } \\ 
	\end{tabular}
	\caption{ \textbf{Left}: Function curve of $f(x) = {e^{tx}}$.  \textbf{Right:} Corresponding disparity probability distribution when increasing $t$. P\_d denotes the disparity probability distribution, and pred is the predicted disparity. Note that we select to directly change $t$ in the inference process to give an intuitive visualization in this toy sample. The $t$ is a fixed hyperparameter during training and inference in our final implementation.}
	\label{fig: t}
\end{figure*}

Here, the temperature parameter $t$ controls the sharpness of the probability distribution. As $t$ increases, the disparity distribution becomes more peaked, encouraging the network to focus on the correct disparity. Fig.~\ref{fig: t} shows a toy sample employing anisotropic soft argmin. As shown, we can simply increase $t$ to consistently improve the gradient of $f^{'}(x)$ and the sharpness of distribution, thus generating a more accurate disparity. Let us define the smooth L1 using the proposed temperate soft argmin as $ L^{'}$. The final loss function can be represented as:
\begin{eqnarray}
L^{'} = {L^{'}_{s}}({d^s},\widehat {{d^{s'}}}) + \lambda {L_u}({p^t}).
\end{eqnarray}

By adjusting the temperature parameter $t$, we can modulate the sharpness of the disparity distribution during both training and inference, resulting in more accurate disparity predictions.

\noindent\textbf{Gradient analysis.} In this section, we claim the proposed anisotropic soft argmin can further accelerate the training convergence. As shown in Eq.~\ref{gradient}, the gradient of smooth L1 with anisotropic soft argmin $L^{'}_s$ is t times larger than the original smooth L1 implementation $L_s$, which means that the proposed anisotropic softmax has a greater penalty for false disparity estimation. Hence, we can employ the proposed anisotropic soft argmin to accelerate the training speed and push the network more quickly converging to the right distribution. Meanwhile, $L^{'}_s$ assigns a larger gradient to the wrong prediction, thus constraining the network to focus on these hard examples and correct the prediction.
\begin{eqnarray}
\frac{\partial L_s^{'}}{\partial C(i)} = \frac{\partial L_s^{'}}{\partial p(i)} \frac{\partial p(i)}{\partial C(i)} = t \frac{\partial L_s}{\partial p(i)} \frac{\partial p(i)}{\partial C(i)}
\label{gradient}
\end{eqnarray}

\subsection{Pseudo-Label-based Self-Distillation}

Pseudo-label-based knowledge distillation~\cite{AOHNet, unsuperviseddomainadaptation, shen2023digging} is a common approach for domain adaptation in stereo matching. Inspired by these works, we propose using our uncertainty metric to generate pseudo-labels for further adaptation. Specifically, we set a threshold $t_\delta$ to filter high-uncertainty pixels in the predicted disparity map:
\begin{eqnarray}
d_f = \left\{
\begin{array}{ll}
\hat{d}, & \text{if } U(p^t) < t_\delta, \\
0, & \text{if } U(p^t) > t_\delta,
\end{array}
\right.
\end{eqnarray}
where 0 indicates invalid pixels. The threshold $t_\delta$ filters out the $\delta\%$ of pixels with the highest uncertainty. The generated pseudo-labels are then used to adapt the network to the target domain in a supervised manner. Our experiments show that the proposed \textit{Anisotropic Soft Argmin} and uncertainty estimation complement the pseudo-label generation process, further improving the quality of the pseudo-labels.

\section{Experiments}
\label{Experiment}
\subsection{Datasets}
We employ three publicly available datasets: SceneFlow~\cite{dispnet}, ETH3D \cite{eth3d} and KITTI 2012 \& 2015\cite{KITTI_2012,KITTI_Stereo_2015} to train and evaluate our proposed method. Below we will introduce each dataset for more detail.

\noindent \textbf{SceneFlow~\cite{dispnet}} is a large synthetic dataset containing 35454 training pairs and 4370 testing pairs. It contains FlyingThings3D, Driving, and Monkaa subsets with dense and accurate ground truth for training.

\noindent \textbf{KITTI 2015 \& KITTI 2012:\cite{KITTI_2012,KITTI_Stereo_2015}} are real-world datasets collected from a driving car with sparse ground-truth disparity. Specifically, KITTI 2015 contains 200 training and 200 testing image pairs, while KITTI 2012 provides 194 training and 195 testing image pairs.

\noindent \textbf{ETH3D:\cite{eth3d}}: is a grayscale image
dataset with both indoor and outdoor scenes, which contains 27 training images and 20 testing images. We use it to test the adapt generalization of the proposed method. 







\subsection{Implementation details}
We embed our method into three representative stereo matching networks: GwcNet \cite{gwcnet}, PCWNet\cite{shenpcw}, and ITSA \cite{itsa}, namely GwcNet\_CMD, PCWNet\_CMD and ITSA\_CMD. GwcNet has been well-studied and commonly employed as a baseline in many prior works; ITSA and PCWNet are two recently proposed state-of-the-art stereo matching networks focusing on cross-domain generalization. Hence, we select these models as our baseline model.
Specifically, all networks are implemented by PyTorch and optimized with Adam (${\beta _{\rm{1}}} = 0.9,{\beta _2} = 0.999$). To make a fair comparison, we maintain the default hyper-parameter setting and data augmentation of the corresponding official implementation. Besides, the batch size is fixed to 8 for training on 8 Teslas V100 GPUs, and the max disparity search range is set to 192 for all models. More specifically, our training process can be broken down into three steps. Firstly, we pre-train all networks on scene flow datasets with their default training strategy. Then, the model will be finetuned on combining the SceneFlow dataset and unlabeled target domain data with the proposed loss function for 20 epochs. The initial learning rate is 0.001 and is down-scaled by 10 after epoch 12. $\lambda$ is set to 1, 0.5, 0.125 for PER, MSM, and Entropy metric. Finally, we will employ the best model in step two to generate the corresponding pseudo-label for further domain adaptation. 
\renewcommand\arraystretch{1}
\begin{table}[t!]
\centering
{
\caption{Ablation study on the KITTI 2015 training dataset. $t$ is the hyper-parameter of anisotropic soft argmin. $L_u$ denotes the proposed uncertainty minimization loss and $d_f$ represents the generated pseudo-label. $U_m$ is the selected uncertainty metric, \textit{e.g.}, PER, MSM, and Entropy.}
\resizebox{0.48\textwidth}{!}{
\label{tab:abstudy}
\begin{tabular}{c|c|c}
\toprule
Experiment                               & Method                          & D1\_all(\%)           \\ \midrule
     & $t=1$, no $L_u$                             & 15.8                  \\  
                             Anisotropic             & $t=4$, no $L_u$                             & 11.9                  \\  
               soft argmin                        & $t=16$, no $L_u$                            & \textbf{8.7 }                  \\  
                                          & $t=32$, no $L_u$                            & 13.1                  \\ \midrule
 & $t=1$, no $L_u$                      & 15.8                  \\  
                   Uncertainty-regularized                       & $t=1$, $L_u$, ${U_m} == \text{MSM}$                        & 7.9                 \\ 
            minimization                              & $t=1$, $L_u$, ${U_m} == \text{Entropy}$                    & 6.8                 \\  
                                          & $t=1$, $L_u$, ${U_m} == \text{PER}$                        & \textbf{6.6}                   \\ \midrule
\multirow{3}{*}{Combination}              & $t=16$, $L_u$, ${U_m} == \text{MSM}$                        & 6.6                \\  
                                          & $t=16$, $L_u$, ${U_m} == \text{Entropy}$                    & 6.4               \\  
                                          & $t=16$, $L_u$, ${U_m} == \text{PER}$                        & \textbf{6.1}               \\ \midrule
{Self-distillation}        & \multicolumn{1}{l|}{$t=16$, $L_u$, ${U_m} == \text{PER}$, no $d_f$} & 6.1 \\  
          with $d_f$                                 & \multicolumn{1}{l|}{$t=16$, $L_u$, ${U_m} == \text{PER}$, use $d_f$} & \textbf{4.1} \\ \bottomrule
\end{tabular}
}
}

\end{table}
\renewcommand\arraystretch{1}

\subsection{Ablation study}
We perform various ablation studies to show the relative importance of each component. GwcNet is selected as our baseline model, \textit{i.e.}, $t=1$, no $L_u$. All methods are trained on the Scneflow dataset (source domain) and unlabeled KITTI2015 dataset (target domain). The D1\_all error rate of KITTI2015 is reported, and the results are shown in Tab. \ref{tab:abstudy}. Below we describe each component in more detail.

\textbf{Anisotropic soft argmin.} The temperature $t$ is a hyper-parameter that can control the sharpness of disparity probability distribution, \textit{e.g.}, a higher $t$ leads to a more sharp distribution. In this section, we test the influence of employing different $t$. As shown in Tab. \ref{tab:abstudy}, we can achieve a better generalization performance by increasing $t$, which further validates our hypothesis that the constraint of the target domain's multimodal distribution can lead to better generalization. We also notice that $t$ is unsuitable for infinite growth, and using $t=16$ achieves the best performance, likely because we still need a suitable predominantly unimodal distribution to smooth the output and generate sub-pixel disparity estimates rather than a complete unimodal distribution. Subfigure (b) in Fig. \ref{fig: intro_visual} gives a detailed sample of such a case.

\textbf{Uncertainty minimization.} We test the influence of employing different uncertainty metrics as the loss term. As shown in Tab. \ref{tab:abstudy}, all uncertainty metrics can achieve consistent generalization improvement over baseline methods, which further verifies that the proposed uncertainty minimization is a general solution for domain adaptation and it is insensitive to the selected uncertainty metrics. Besides, we also observe that PER and Entropy metrics can obtain a larger gain than MSM. This is because the former two are global uncertainty measures and can better access the sharpness of distribution than local uncertainty measures, \textit{i.e.}, MSM.

\begin{figure}[!t]
	\centering
	\tabcolsep=0.05cm
    \includegraphics[width=0.85\linewidth]{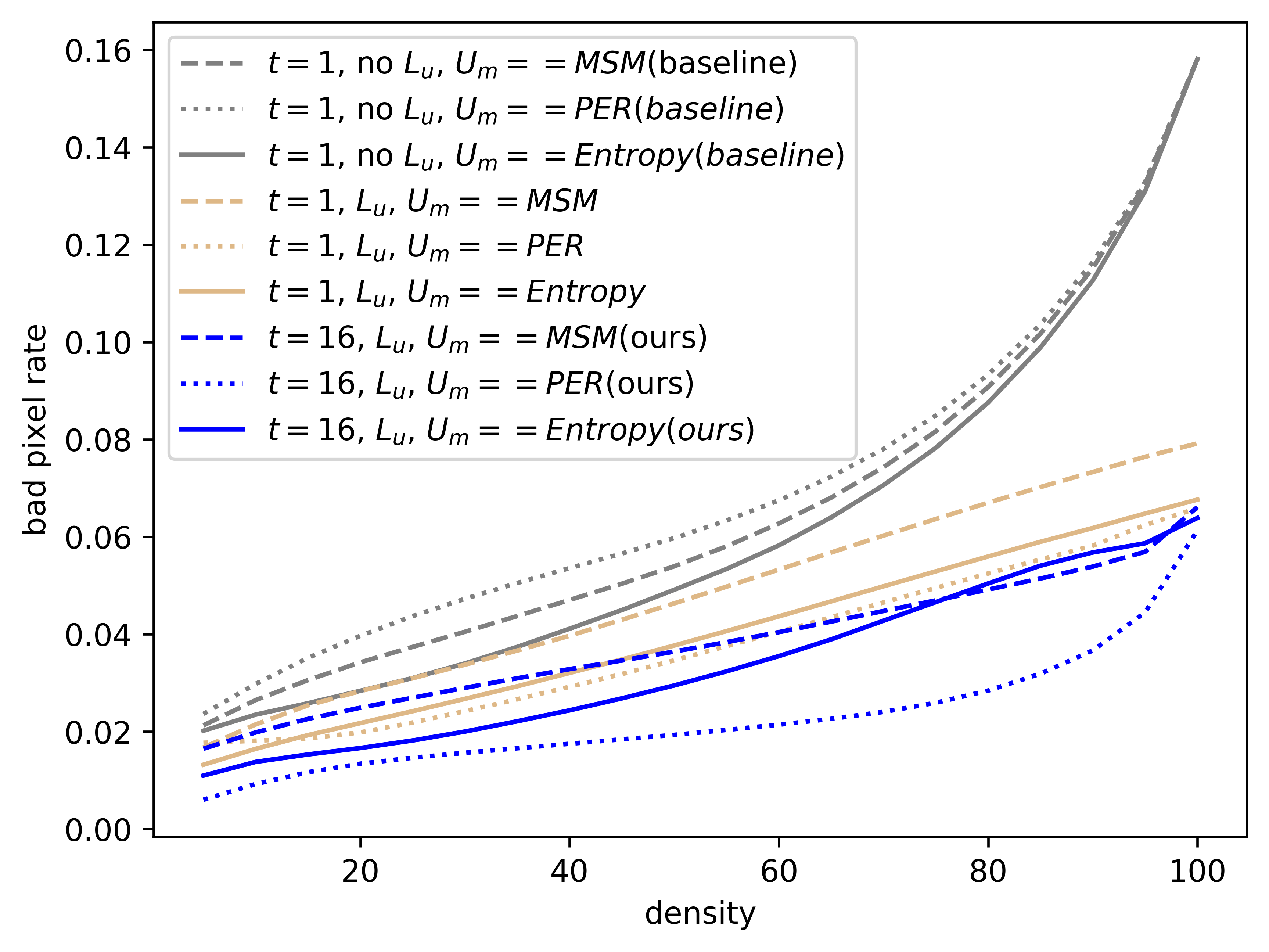}
    \caption{The ROC curves of the proposed method on KITTI2015 dataset. D1\_all is used for evaluation (the lower, the better). $L_u$ denotes the proposed uncertainty minimization loss, $U_m$ is the selected uncertainty metric, and $t$ is the hyper-parameter of anisotropic soft argmin. ($t=1$, no $L_u$) denotes the baseline setting and ($t=16$, $L_u$) is our method. Density denotes the valid pixel in the predicted pseudo-label.}
\label{fig: roc}
\end{figure}

\textbf{Combination impact.} We test the impact of combining the proposed anisotropic soft argmin and uncertainty-regularized minimization. As shown in Tab.\ref{tab:abstudy}, the proposed two methods complement each other and can further boost performance. Specifically, the D1\_all error rate is decreased from 8.7\% ($t=16$, no $L_u$) to 6.1\% ($t=16$, $L_u$, ${U_m} == PER$) after adding PER uncertainty minimization to the best model in anisotropic soft argmin. Similar gains can be observed on other uncertainty metrics.

\textbf{Self-distillation with $d_f$.} The proposed method is also beneficial to the pseudo-label generation. Inspired by the standard evaluation metric in confidence estimation, we employ the ROC curve \cite{confidencesurvey} to quantitatively evaluate our results. Specifically, we first sort pixels in the predicted disparity map following a decreasing uncertainty order. Then, we compute the D1\_all error rate on sparse maps obtained by iterative filtering (\textit{e.g.}, 5\% of pixels each time) from the dense map and plot the ROC curve. See the corresponding result in Fig. \ref{fig: roc}. As shown, the proposed uncertainty minimization (yellow line) can dramatically decrease the error rate of filtered disparity maps (gray line) at any density with all uncertainty metrics, and the combination of the proposed two methods (blue line) can achieve a larger gain, in which the model using PER uncertainty minimization and anisotropic soft argmin ($t=16$, $L_u$, ${U_m} == PER$) obtains the best performance. Hence, we propose to employ this model to generate pseudo-labels for further domain adaptation. Results in Tab. \ref{tab:abstudy} show that we can further decrease the D1\_all error rate from 6.1\% to 4.1\% after adding the supervision of generated pseudo-labels $d_f$.

 \begin{figure*}[!t]
 	\centering
 	\small
 	\tabcolsep=0.05cm
 	\begin{tabular}{c c c c c}
 	
 	\includegraphics[width=0.19\linewidth]{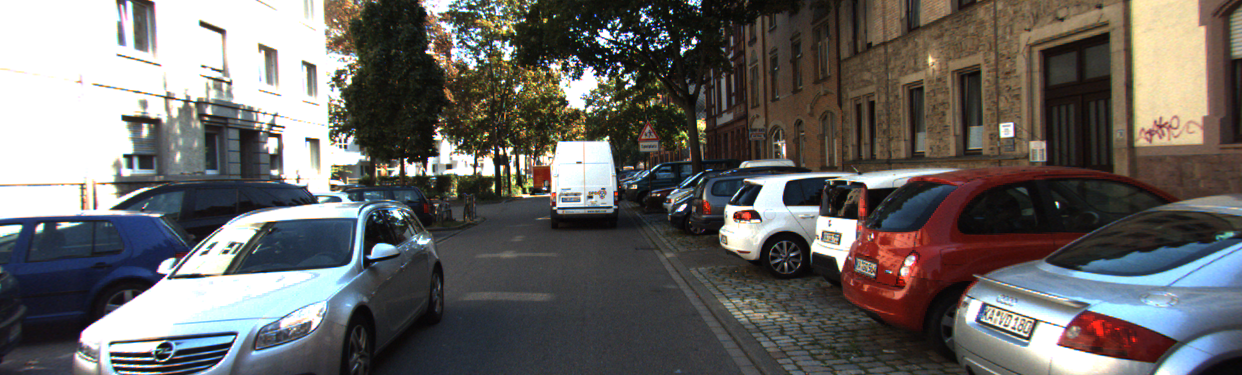}&
 	\includegraphics[width=0.19\linewidth]{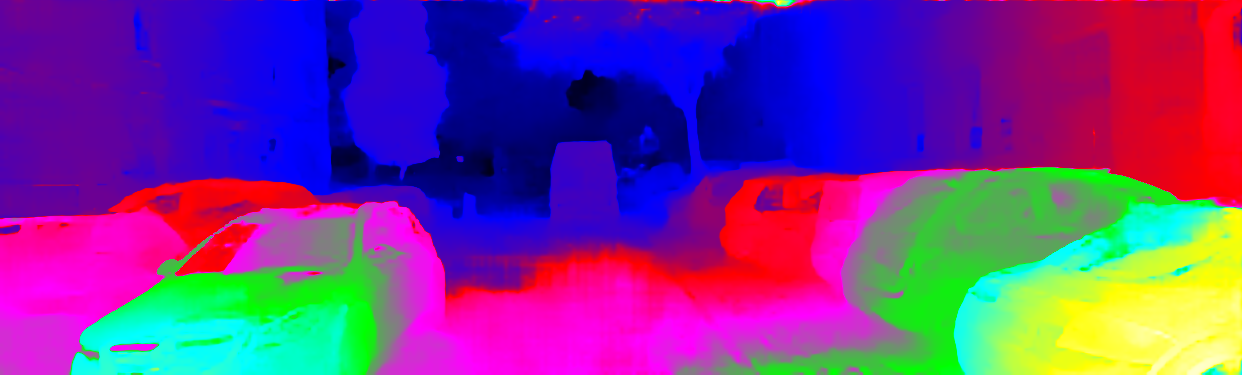}&
 	\includegraphics[width=0.19\linewidth]{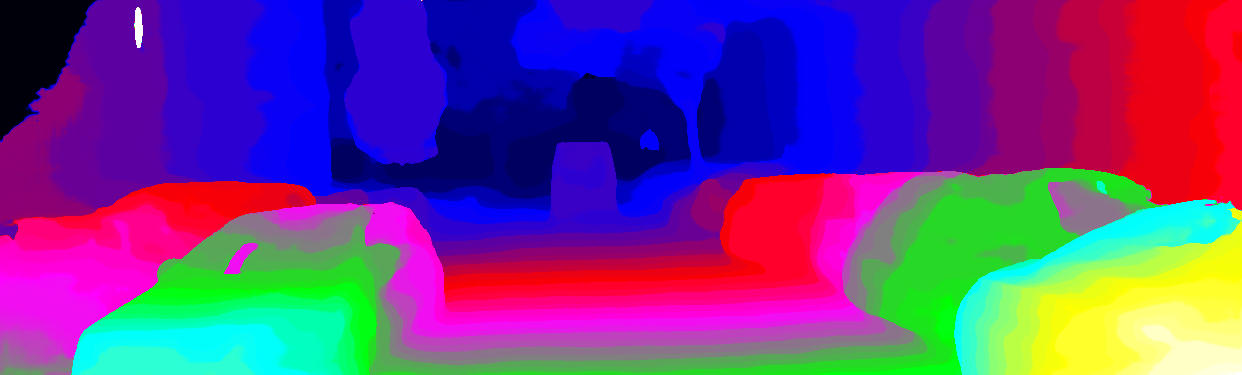}&
 	\includegraphics[width=0.19\linewidth]{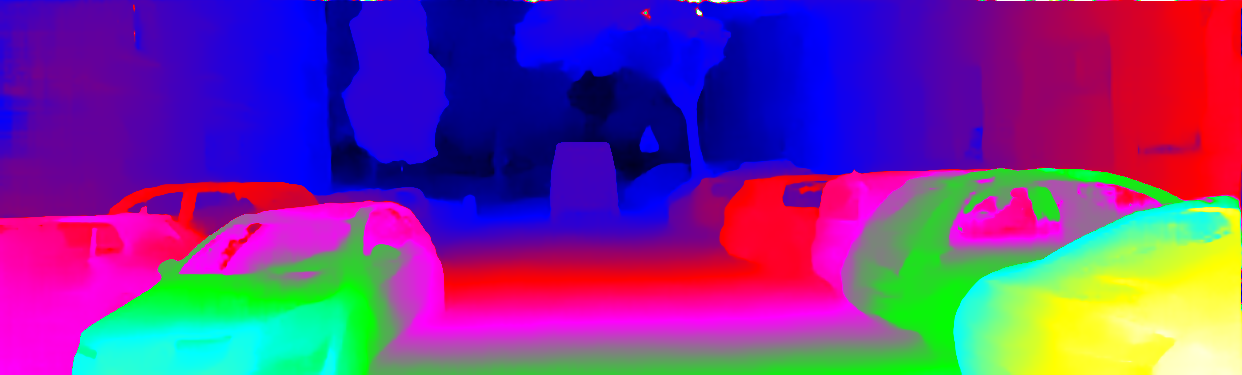}&
        \includegraphics[width=0.19\linewidth]{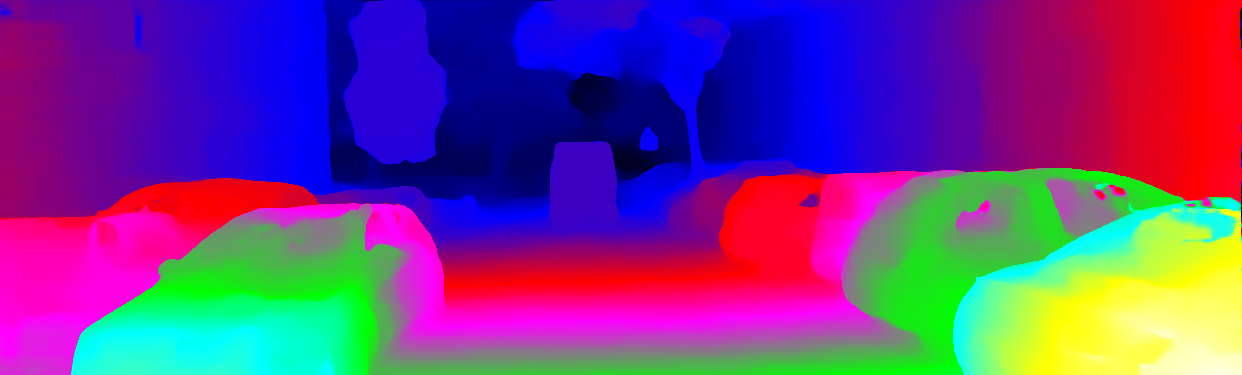}\\
	
 	& \includegraphics[width=0.19\linewidth]{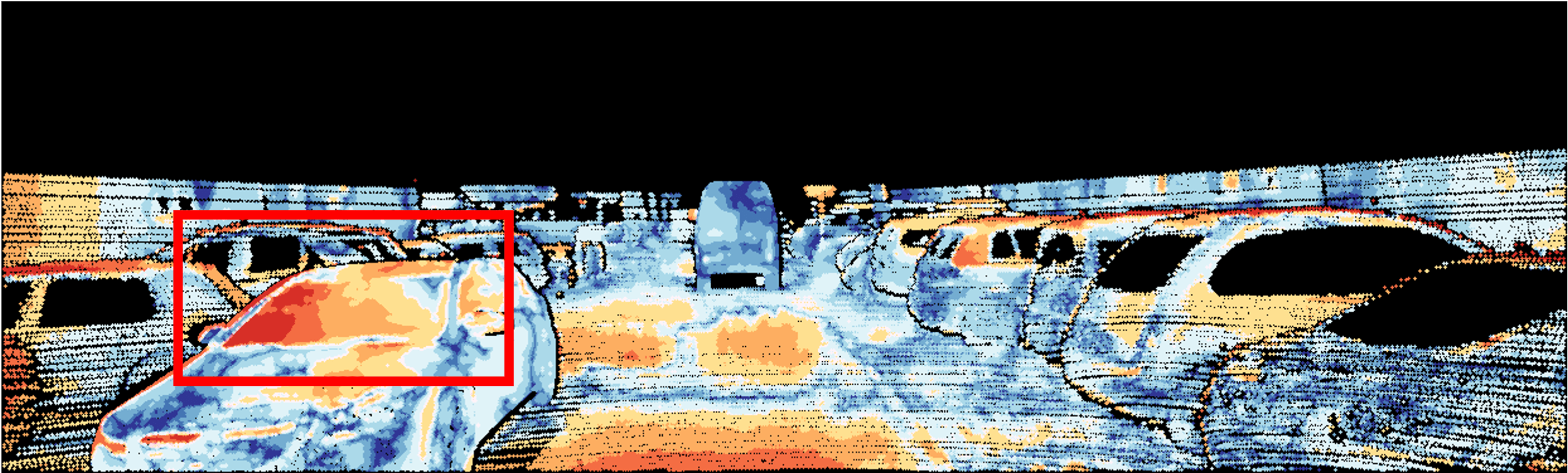}&
 	\includegraphics[width=0.19\linewidth]{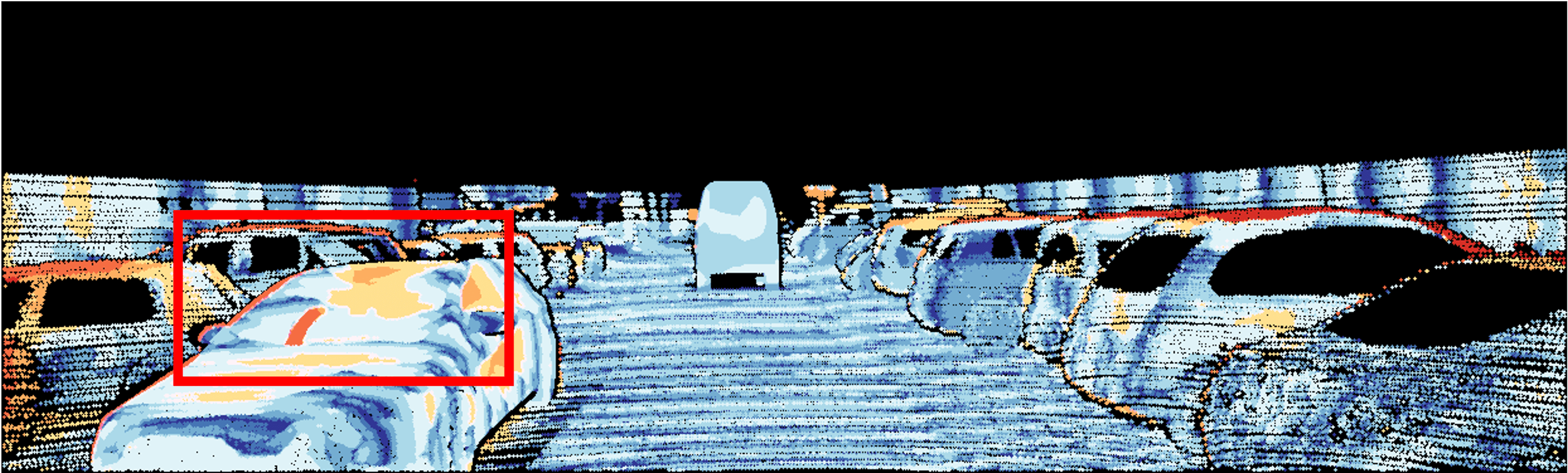}&
 	\includegraphics[width=0.19\linewidth]{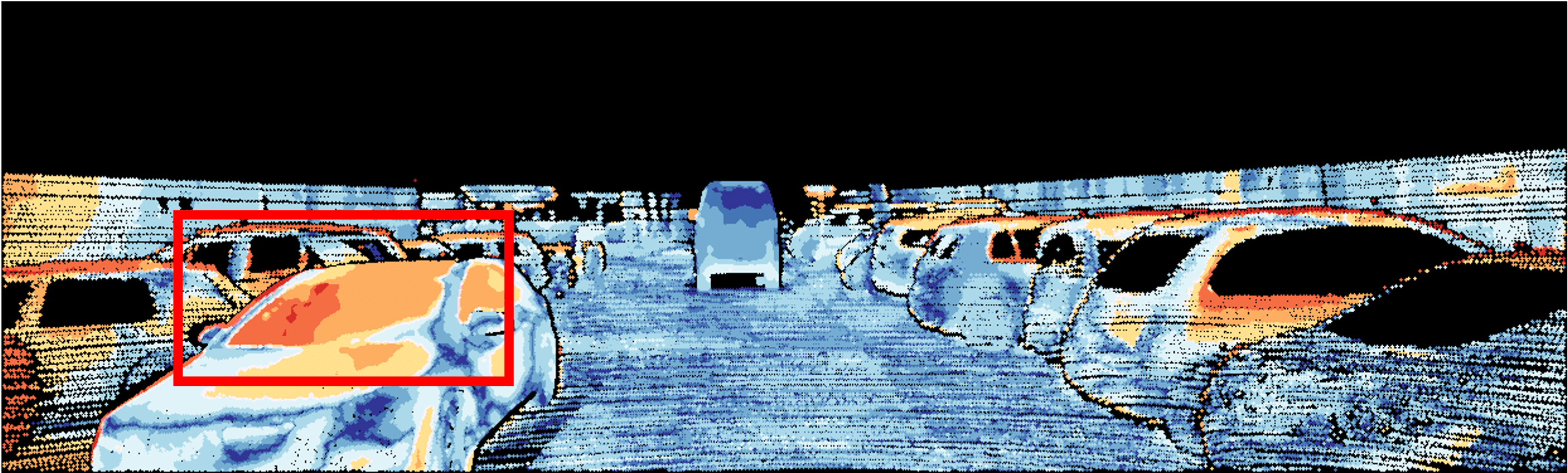}&
        \includegraphics[width=0.19\linewidth]{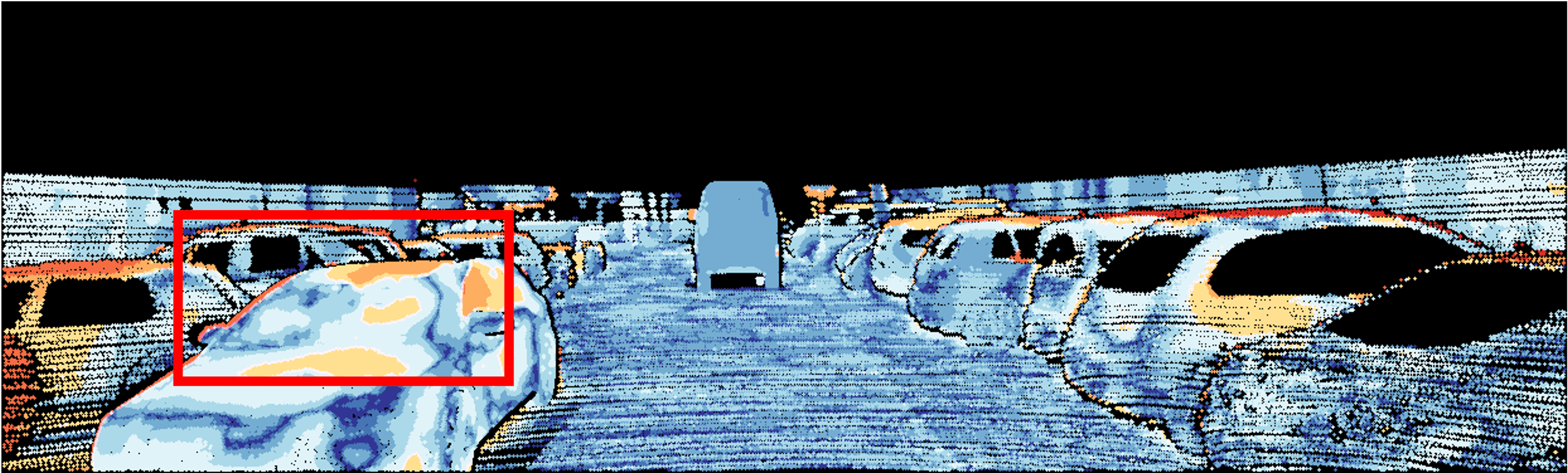}\\

   	\includegraphics[width=0.19\linewidth]{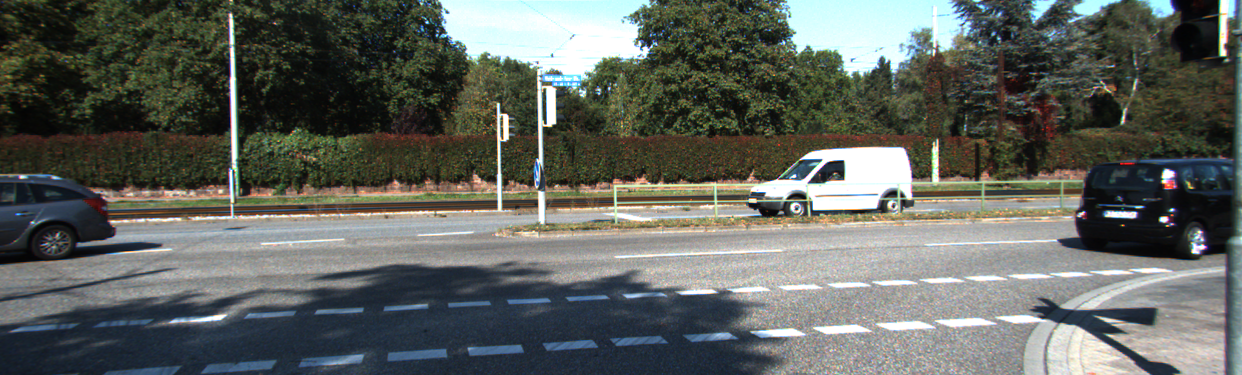}&
 	\includegraphics[width=0.19\linewidth]{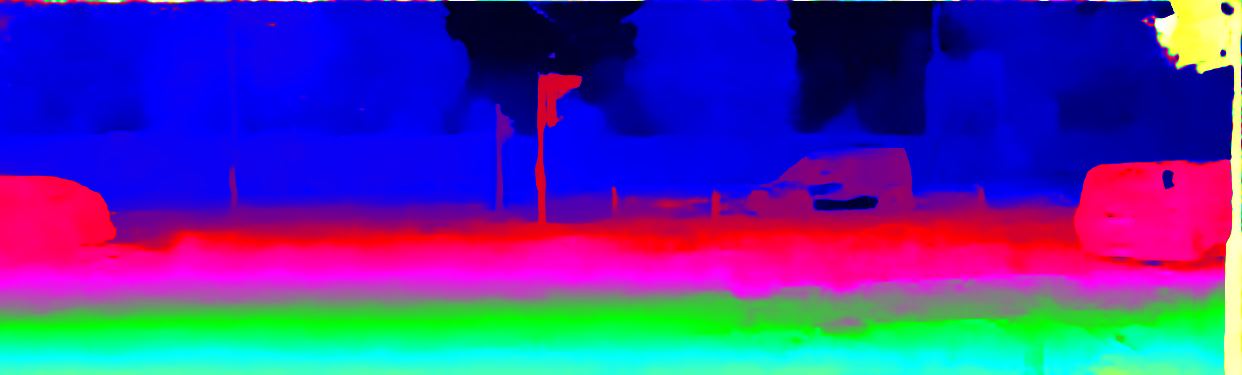}&
 	\includegraphics[width=0.19\linewidth]{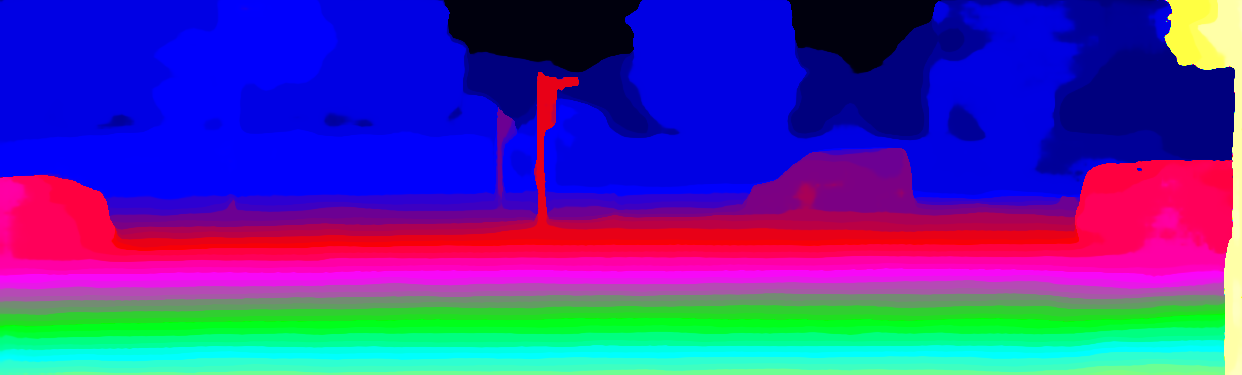}&
 	\includegraphics[width=0.19\linewidth]{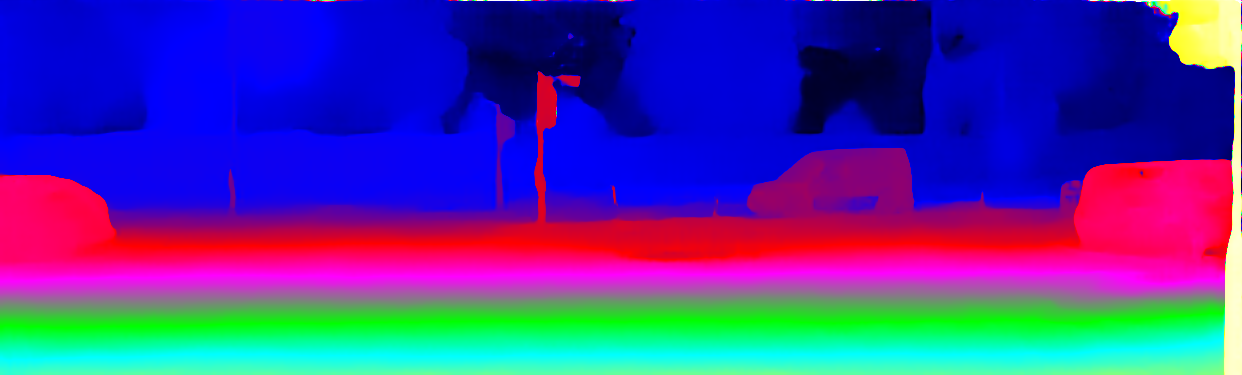}&
        \includegraphics[width=0.19\linewidth]{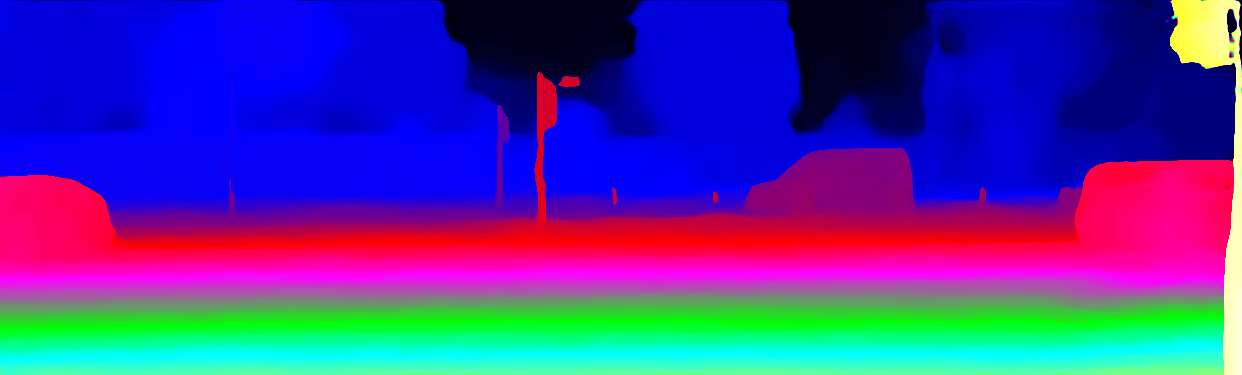}\\
	
 	& \includegraphics[width=0.19\linewidth]{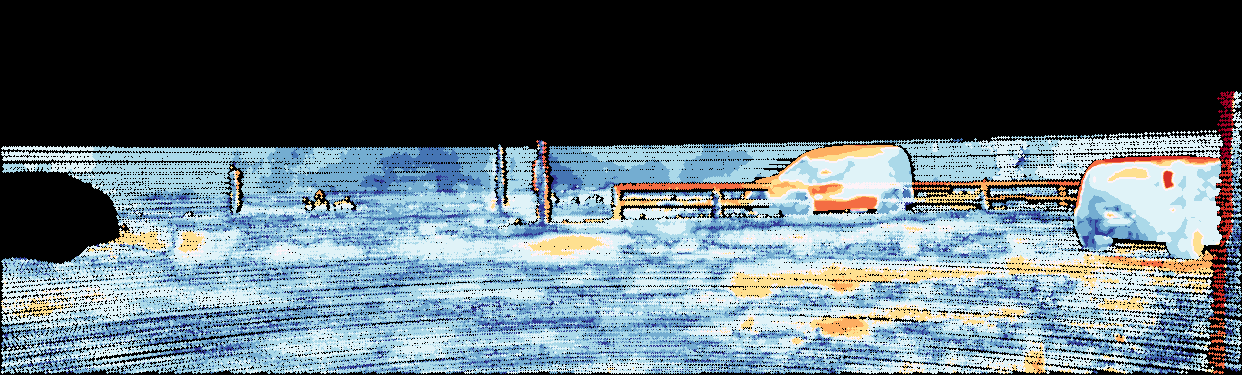}&
 	\includegraphics[width=0.19\linewidth]{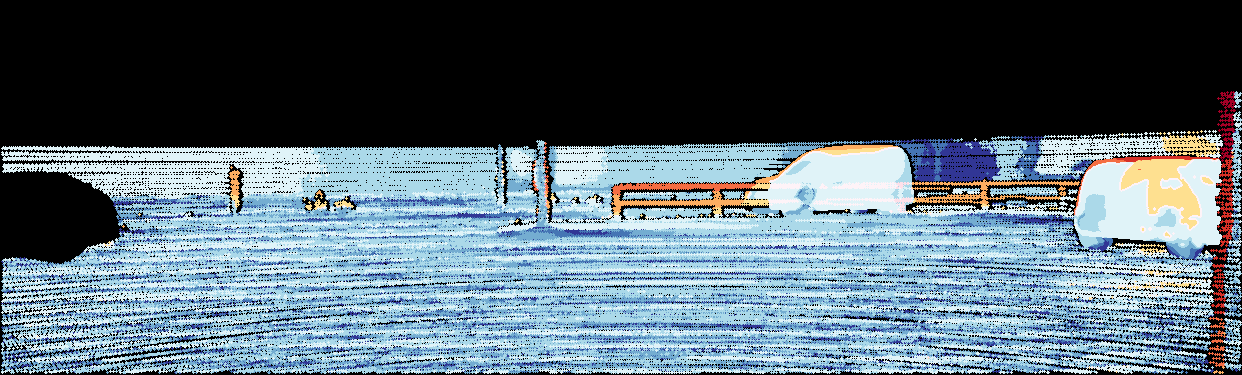}&
 	\includegraphics[width=0.19\linewidth]{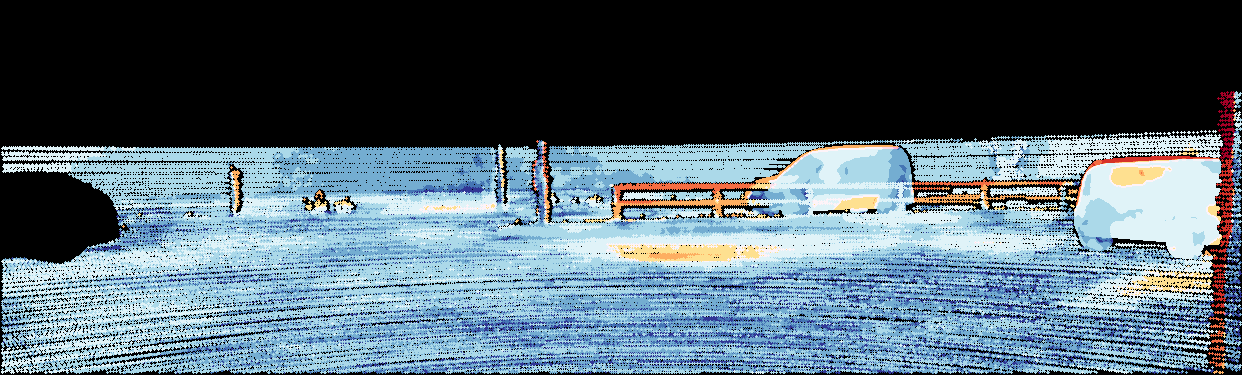}&
        \includegraphics[width=0.19\linewidth]{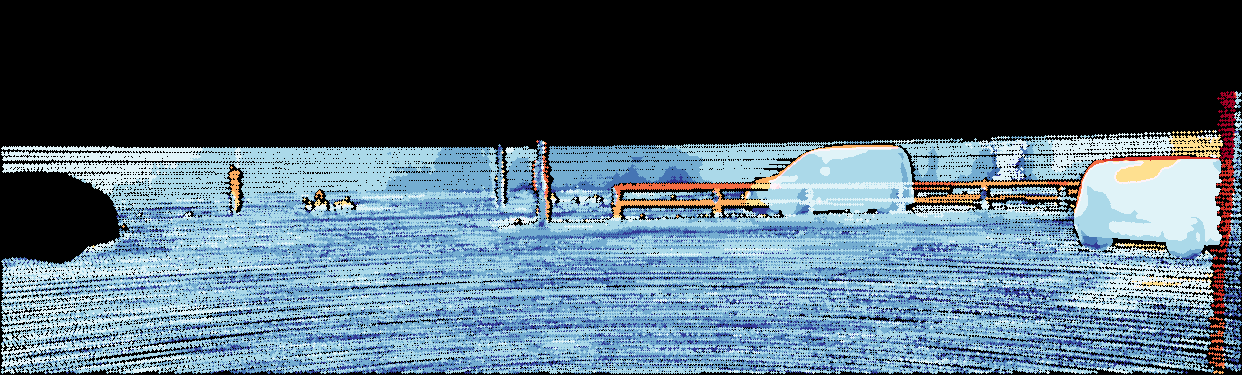}\\

 	{(a) Left image} &	{ (b) GWCNet }	&  { (c) GWCNet\_CMD}	&	{ (d) ITSA}  & { (e) ITSA\_CMD}	\\
 	\end{tabular}
 	\caption{Comparisons of cross-domain generalization on KITTI2015 trainset. The first column shows the input left images, and for each following column, the top row shows the predicted colorized disparity map and the bottom row shows the error map.}
 	\label{fig: generalization}
 \end{figure*}
 
\begin{figure}[!t]
	\centering
	\tabcolsep=0.02cm
	\begin{tabular}{c c}
	\includegraphics[width=0.45\linewidth]{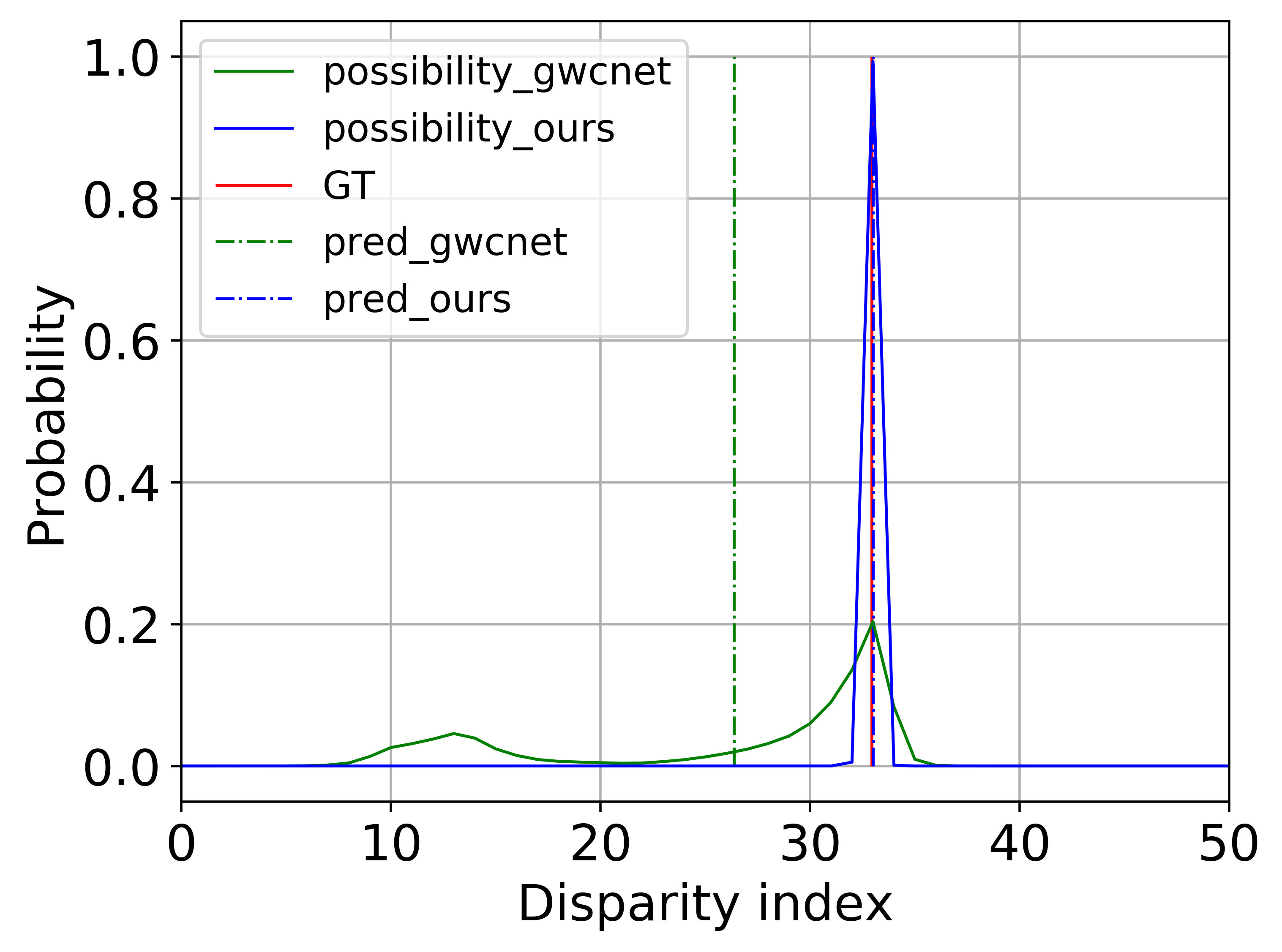}&
     \includegraphics[width=0.45\linewidth]{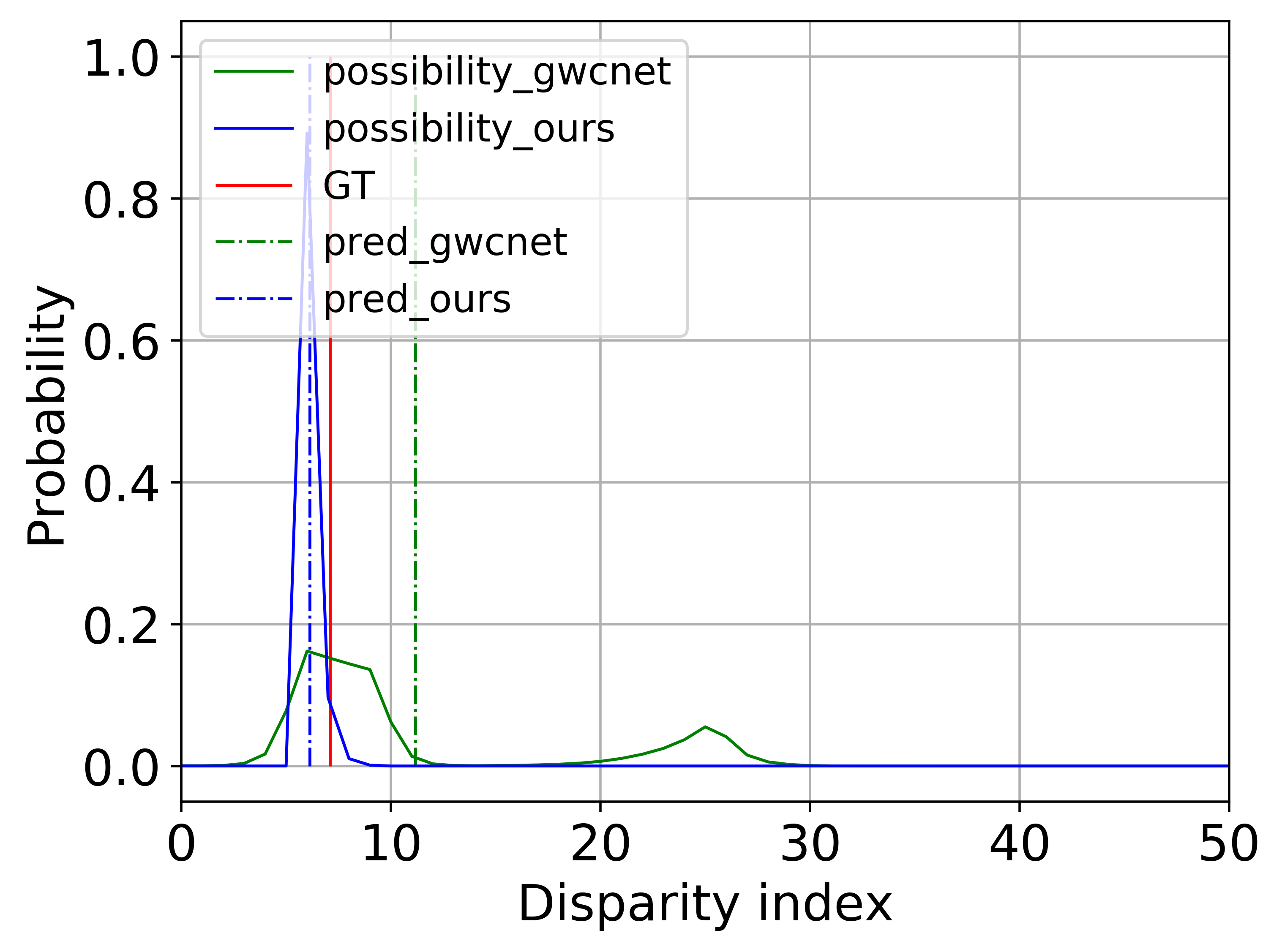}\\
	 \multicolumn{2}{c}{(a)} \\
 	\includegraphics[width=0.45\linewidth]{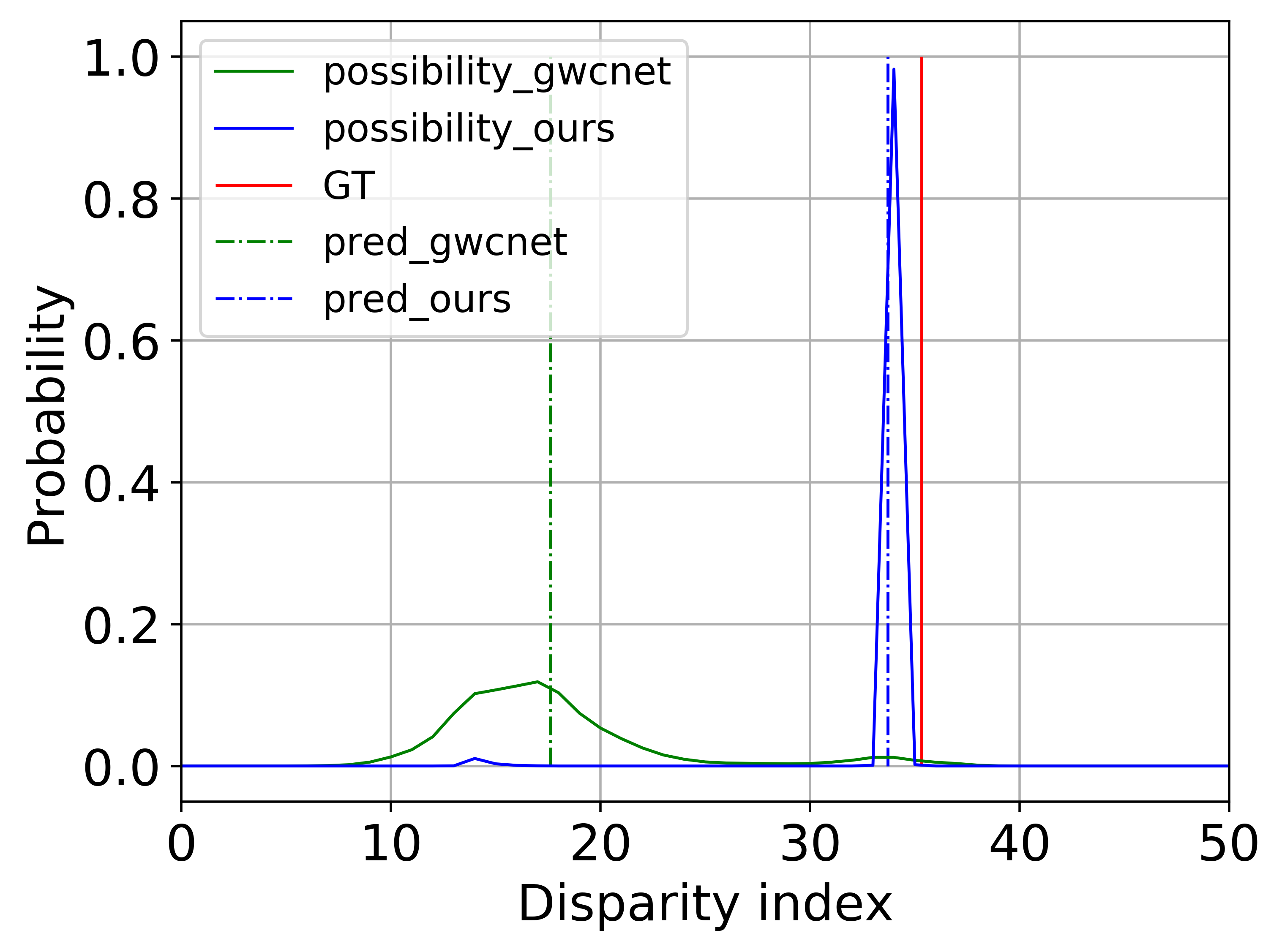}&
     \includegraphics[width=0.45\linewidth]{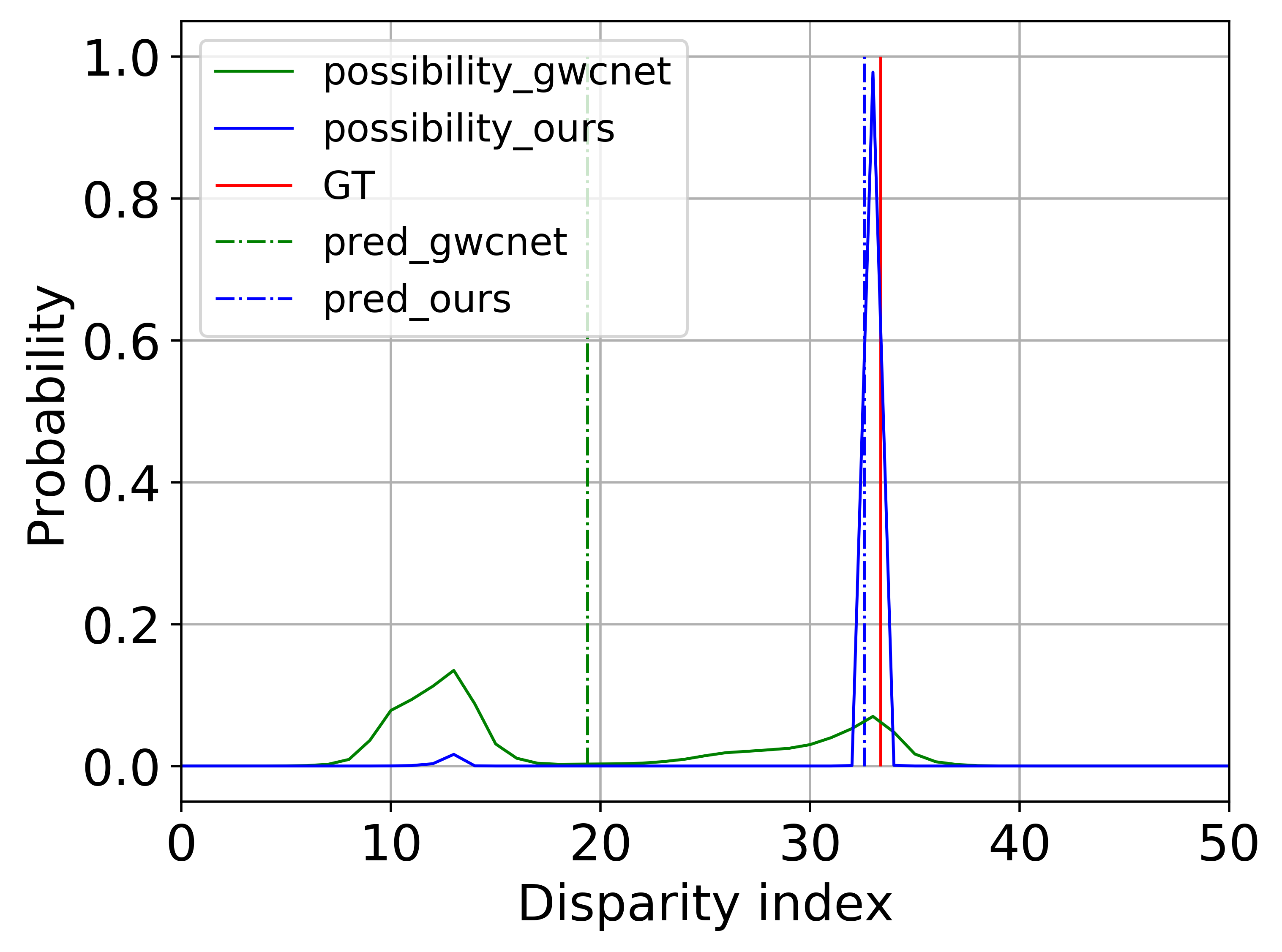}\\
     \multicolumn{2}{c}{(b)} \\
	\end{tabular}
 \vspace{-0.1in}
	\caption{More detailed disparity probability distribution comparison between GWCNet and GWCNet\_CMD. All methods are trained on the combination of the SceneFlow dataset and
unlabeled target domain data. Two wrong disparity probability distributions can be corrected by the proposed method: (a) method predicts a multimodal distribution while it can predict a higher probability at the true disparity index. (b) method directly predicts the highest probability at the wrong disparity index while it still can predict some faint probability at the true disparity index.
}
	\label{fig: distribution}
\end{figure}

\textbf{Comparison with argmin.} \B{Recall our goal is to restrict the multimodal disparity distributions of the target domain. Using the argmin operation during inference is the most straightforward approach, as it directly selects the disparity index with the highest probability to address the multimodal distribution issue. Here, we test such impact in Tab. \ref{tab:argmin}. As shown, the performance gain of GWCNet\_argmin compared to GWCNet is limited, which means such a way can't effectively solve the multimodal distributions issue.} This is primarily due to the argmin operation can only optimize examples that can predict a higher probability at the true disparity index, and it will even degrade performance when the network indeed predicts a correct disparity from a predominantly unimodal distribution by soft argmin. Instead, the proposed method can further decrease the error rate from 14.1\% to 6.1\% compared to GWCNet\_argmin, a 56.74\% error rate reduction. This is mainly because our method can go beyond argmin and directly correct examples that directly predict the highest probability at the wrong disparity index by pushing the distribution converging to the right distribution, \textit{i.e.}, a unimodal distribution peaked at the ground truth disparity. Visualization can be seen in Fig. \ref{fig: distribution}(b) and see a detailed explanation in the next section.

\begin{table}[t!]
\caption{Comparison with using argmin on the KITTI dataset.}
\centering
\resizebox{0.25\textwidth}{!}{
\begin{tabular}{c|c}
\toprule
Method                          & D1\_all(\%)           \\ \midrule
      GWCNet                            & 15.8                  \\  
                                           GWCNet\_argmin                         & 14.1                  \\  
                                           GWCNet\_CMD                            & \textbf{6.1}                  \\  \bottomrule
\end{tabular}
}
\label{tab:argmin}
\end{table}

\subsection{Disparity probability distribution analysis}
In this section, we provide a comprehensive analysis of the disparity probability distribution that the proposed method aims to optimize. All models are trained using a combination of the SceneFlow dataset and unlabeled data from the target domain. As shown in Fig. \ref{fig: distribution}, the proposed method can alleviate two wrong disparity probability distributions: (1) Method predicts a multimodal distribution while it can predict a higher probability at the true disparity index (Fig. \ref{fig: distribution}-(a)). For this case, we can directly constrain the multimodal distribution of networks to obtain a more accurate disparity prediction result. (2) The method directly predicts the highest probability at the wrong disparity index while it still can predict some faint probability at the true disparity index (Fig. \ref{fig: distribution}-(b)). For this case, we mainly rely on the stronger penalty of anisotropic soft argmin to push the distribution converging to the right distribution, \textit{i.e.}, an unimodal distribution peaked at the ground truth disparity. 
Both quantitative (Tab.~\ref{tab:sharpness}\&~\ref{tab:constraint}) and qualitative results demonstrate that the proposed method successfully alleviates the issue of multimodal distribution on the target domain and achieves better generalization.
\subsection{Comparison with constraint on source domain}
We also notice some work \cite{acfnet, pds} observes the multi-modal distribution issue and 
explores to directly constrain the multi-modal distribution in the source domain.
Specifically, PDSNet\cite{pds} and AcfNet\cite{acfnet} propose to construct a ground truth disparity probability distribution and supervise the network with cross-entropy loss. However, these methods can not alleviate the multimodal distribution issues in the target domain. A direct comparison experiment is employed in Tab. \ref{tab:constraint} to support our claim. To make a fair comparison, we add the cross-entropy loss to the same baseline model GWCNet, namely GWCNet\_PDS. As shown, GWCNet\_PDS cannot constrain the multimodal distribution in the target domain, and the performance gain for generalization is limited. Moreover, due to the absence of ground truth disparity, we \textit{cannot} directly apply such a method to the target domain in the domain adaptation setting. Instead, our method can greatly decrease the uncertainty of target domains and further decrease the error rate from 13.2\% to 6.1\% compared to GWCNet\_PDS, a 53.7\% error rate reduction. Additionally, we can observe that the uncertainty metric has a consistent decline, \textit{i.e.}, the MSM metric has decreased from 0.73 to 0.32. This phenomenon further verifies our claim that we can force the network to output a predominantly unimodal distribution on the target domain for better generalization.

\begin{table}[t!]
\caption{Comparison with Constraint on Source Domain. All methods are trained on the combination of SceneFlow dataset and unlabeled target domain data. MSM, PER, and Entropy are used to evaluate the sharpness of distribution.}
\centering
\resizebox{0.45\textwidth}{!}{
\begin{tabular}{c|c|c|c|c}
\toprule
Method                          &  D1\_all(\%)   & MSM  & Entropy & PER          \\ \midrule
      GWCNet \cite{gwcnet}                            & 15.8     & 0.72 & 1.96 & 0.92             \\  
      GWCNet\_PDS   \cite{pds}     & 13.2    & 0.73 & 2.08  &0.92               \\  
    GWCNet\_CMD      & \textbf{6.1} & \textbf{0.32} & \textbf{0.46} & \textbf{0.64}         \\  \bottomrule
\end{tabular}
}
\label{tab:constraint}
\end{table}

\begin{table}[t!]
\centering
{
\caption{Domain adaptation evaluation on KITTI 2012 and KITTI 2015 training dataset. `*
means adaptation on test sets of the target domain and evaluation on training sets and `$\dagger$' means both adaptation and evaluation on the training set of the target domain. All the values are the lower the better.}
\resizebox{0.4\textwidth}{!}{ 
\label{tab:itsa&gwc}
\begin{tabular}{l|c|c}
\toprule
Method    & \begin{tabular}[c]{@{}c@{}}KITTI 2012\\ D1\_all (\%)\end{tabular} & \begin{tabular}[c]{@{}c@{}}KITTI 2015\\ D1\_all (\%)\end{tabular}   \\ \midrule
GWCNet \cite{gwcnet}   & 14.9   & 15.8                      \\ 
GWCNet\_CMD*  & \textbf{3.6}   & 4.3                      \\   
GWCNet\_CMD\textsuperscript{$\dagger$}   & 3.8  & \textbf{4.1}                      \\  \midrule
ITSA \cite{itsa}   & 4.8   & 5.9                      \\
ITSA\_CMD*                 & \textbf{3.2}                    & 3.8        \\
ITSA\_CMD\textsuperscript{$\dagger$}  & 3.4   & \textbf{3.6}                     \\ \midrule
PCWNet \cite{shenpcw}   & 4.2   & 5.6                      \\ 
PCWNet\_CMD*                 & \textbf{3.1}                    & 4.0        \\
PCWNet\_CMD\textsuperscript{$\dagger$}  & 3.3   & \textbf{3.7}                     \\ \bottomrule
\end{tabular}
}
}
\end{table}

\subsection{Experiment on more methods}
In this section, we embed our method into three representative stereo models, \textit{i.e.}, GWCNet, PCWNet, and ITSA to further validate the effectiveness of the proposed method. To make a fair comparison, we retrained ourselves on all methods with the official implementation. Inspired by previous work \cite{AOHNet, adastereo_ijcv}, we employ two settings to test the generalization of the proposed method. Specifically, we use `*' to indicate adaptation on test sets of the target domain and evaluation on training sets and $\dagger$ means both adaptation and evaluation on the training set of the target domain. As shown in Tab. \ref{tab:itsa&gwc},  the proposed method can consistently improve the generalization of top-performance and domain-generalized stereo matching networks. Take the current state-of-the-art domain-generalized stereo matching network ITSA as an example. As shown, ITSA\_CMD can obtain noticeable gains in both two settings. Moreover, the no overlap data version ITSA\_CMD* can even achieve better performance than overlap data version ITSA\_CMD\textsuperscript{$\dagger$} in KITTI2012 datasets, which indicates the proposed method can indeed learn the distribution of the target domain rather than solely overfitting on the target data.  A similar situation can also be observed on the other two baselines. See the visualization comparison between our method and the original implementation. (\textit{e.g.}, gwcnet vs gwcnet\_CMD) in Fig. \ref{fig: generalization}.


\begin{table}[!htb]
\centering
{
\caption{Comparison with domain generalization/adaptation methods on KITTI2015, KITTI2012 and ETH3D datasets. `*' means adaptation on test sets of the target domain and evaluation on training sets and `$\dagger$' means both adaptation and evaluation on the training set of the target domain. We highlight the best result in each column in \textbf{bold} and the second-best result in \B{blue}.}
\resizebox{0.47\textwidth}{!}{ 
\label{tab: cross-domain generalization}
\begin{tabular}{l|c|c|c}
\toprule
Method     & \begin{tabular}[c]{@{}c@{}}KITTI2012\\ D1\_all(\%)\end{tabular} & \begin{tabular}[c]{@{}c@{}}KITTI2015\\ D1\_all(\%)\end{tabular} & \begin{tabular}[c]{@{}c@{}}ETH3D\\ bad 1.0(\%)\end{tabular} \\ \midrule
\multicolumn{4}{c}{\emph{Cross-domain Generalization Evaluation}      }                                                                                                                                                                                                                                                  \\ \midrule
GWCNet \cite{gwcnet}                 & 12.0                                                            & 12.2                                                                                                                         & 11.0                                                        \\ 
mask-CFNet \cite{masked}                   & 4.8                                                             & 5.8                                                                                                                & 5.7                                                         \\
CFNet\cite{cfnet}                   & 4.7                                                    & 5.8                                                                                                                 & 5.8                                                \\
PCWNet\cite{cfnet}                   & 4.2                                                    & 5.6                                                                                                                 & 5.2                                                \\
CREStereo++\cite{jing2023uncertainty}                    & 4.7                                                             & 5.2                                                                                                                & 4.4                                                         \\ \midrule
\multicolumn{4}{c}{\emph{Adaptation Generalization Evaluation}      }                                                                                                                                                                                                                                                    \\ \midrule     
Stereogan\textsuperscript{$\dagger$} \cite{stereogan}       & -                                                             & 5.74  & - \\
AdaStereo\textsuperscript{$\dagger$} \cite{adastereo_ijcv}       & \B{\textbf{3.6}}                                                             & \textbf{3.5}                                                                                                                             & \textbf{4.1}                                                           \\
PCWNet\_CMD\textsuperscript{$\dagger$}   & \textbf{3.3}   & \B{\textbf{3.7}}   & \textbf{4.1}                  \\ \midrule 
AOHNet* \cite{AOHNet}        & 8.6                                                             & 7.8                                                                                                                            & -                                                           \\ 
MADNet* \cite{madnet}       & 9.3                                                             & 8.5                                                                                                                            & - \\
PCWNet\_CMD*   & \textbf{3.1}   & \textbf{4.0}   & \textbf{4.5}                  \\ \bottomrule
\end{tabular}
}
}
\end{table}

\subsection{Cross-domain generalization evaluation}
This section compares the proposed best-performance method PCW\_CMD with other domain generalization/adaptation methods on more datasets. Similar to Tab. \ref{tab:itsa&gwc}, we also use two settings for evaluation. As shown in Tab. \ref{tab: cross-domain generalization}, the proposed PCW\_CMD set a new SOTA performance in the domain adaptation setting. Specifically, we can easily find that the proposed method achieves two first-place and 1 second-place among all three datasets in the overlap data division setting(*) and three first-place in no overlap data division setting ($\dagger$). Moreover, compared to the baseline model PCWNet, the error rate of PCWNet\_CMD* on KITTI 2012, KITTI 2015, and ETH3D has been decreased by 26.19\%, 28.57\%, 13.46\%, which further verifies the effectiveness of the proposed method.

\section{Conclusion}

In this work, we introduced a novel framework for improving the generalization of stereo-matching networks through the mitigation of multimodal disparity distributions in target domains. By quantitatively analyzing the disparity probability distributions, we confirmed that existing deep stereo-matching methods often produce multimodal distributions when applied to unseen target domains, which significantly degrades performance. To address this, we proposed two complementary strategies: \textit{Uncertainty-Regularized Minimization} and \textit{Anisotropic Soft Argmin}. These methods effectively encourage the network to output unimodal, sharp disparity distributions, which are crucial for robust generalization in cross-domain settings.
Our experimental results across multiple state-of-the-art stereo-matching architectures demonstrate that the proposed techniques consistently improve performance in challenging domain adaptation scenarios. Moreover, the simplicity and effectiveness of our approach make it a valuable addition to existing stereo-matching pipelines, particularly in applications where labeled data from target domains is scarce or unavailable. \B{However, our method still requires unsupervised domain adaptation in the target domain. In practice, users often prefer models that can be deployed directly in their environments without additional fine-tuning. In future work, we plan to explore how to leverage the uncertainty mechanism to further improve the cross-domain generalization of networks, enabling more effective out-of-the-box performance in unseen domains.}

{
    \small
    \bibliographystyle{ieeenat_fullname}
    \bibliography{main}
}

\end{document}


\title{CMD: Constraining Multimodal Distribution for \\ Domain Adaptation in Stereo Matching -- Supplementary Material} 

\titlerunning{Abbreviated paper title}

\author{First Author\inst{1}\orcidlink{0000-1111-2222-3333} \and
Second Author\inst{2,3}\orcidlink{1111-2222-3333-4444} \and
Third Author\inst{3}\orcidlink{2222--3333-4444-5555}}

\authorrunning{F.~Author et al.}

\institute{Princeton University, Princeton NJ 08544, USA \and
Springer Heidelberg, Tiergartenstr.~17, 69121 Heidelberg, Germany
\email{lncs@springer.com}\\
\url{http://www.springer.com/gp/computer-science/lncs} \and
ABC Institute, Rupert-Karls-University Heidelberg, Heidelberg, Germany\\
\email{\{abc,lncs\}@uni-heidelberg.de}}

\maketitle

\begin{abstract}
In this supplementary material, we provide more details and visualization analysis of the proposed method. Firstly, we introduce the detailed dataset information. Then, disparity probability distribution analysis is provided to further verify the effectiveness of the proposed multi-modal distribution constraint. Finally, we furnish a proof demonstrating the validity of the gradient analysis formula described in the main paper.
\end{abstract}

\section{Datasets}
We employ three publicly available datasets: SceneFlow, KITTI 2012 \& 2015, and ETH3D to train and evaluate our method. Below we will introduce each dataset for more detail.

\noindent \textbf{SceneFlow~\cite{dispnet}} is a large synthetic dataset containing 35454 training pairs and 4370 testing pairs. It contains FlyingThings3D, Driving, and Monkaa subsets with dense and accurate ground truth for training.

\noindent \textbf{KITTI 2015 \& KITTI 2012:\cite{KITTI_2012,KITTI_Stereo_2015}} are real-world datasets collected from a driving car with sparse ground-truth disparity. Specifically, KITTI 2015 contains 200 training and 200 testing image pairs, while KITTI 2012 provides 194 training and 195 testing image pairs.

\noindent \textbf{ETH3D:\cite{eth3d}}: is a grayscale image
dataset with both indoor and outdoor scenes, which contains 27 training images and 20 testing images. We use it to test the adapt generalization of the proposed method. 

\section{Disparity Probability Distribution Analysis}
\begin{figure}[!t]
	\centering
	\tabcolsep=0.02cm
	\begin{tabular}{c c}
	\includegraphics[width=0.4\linewidth]{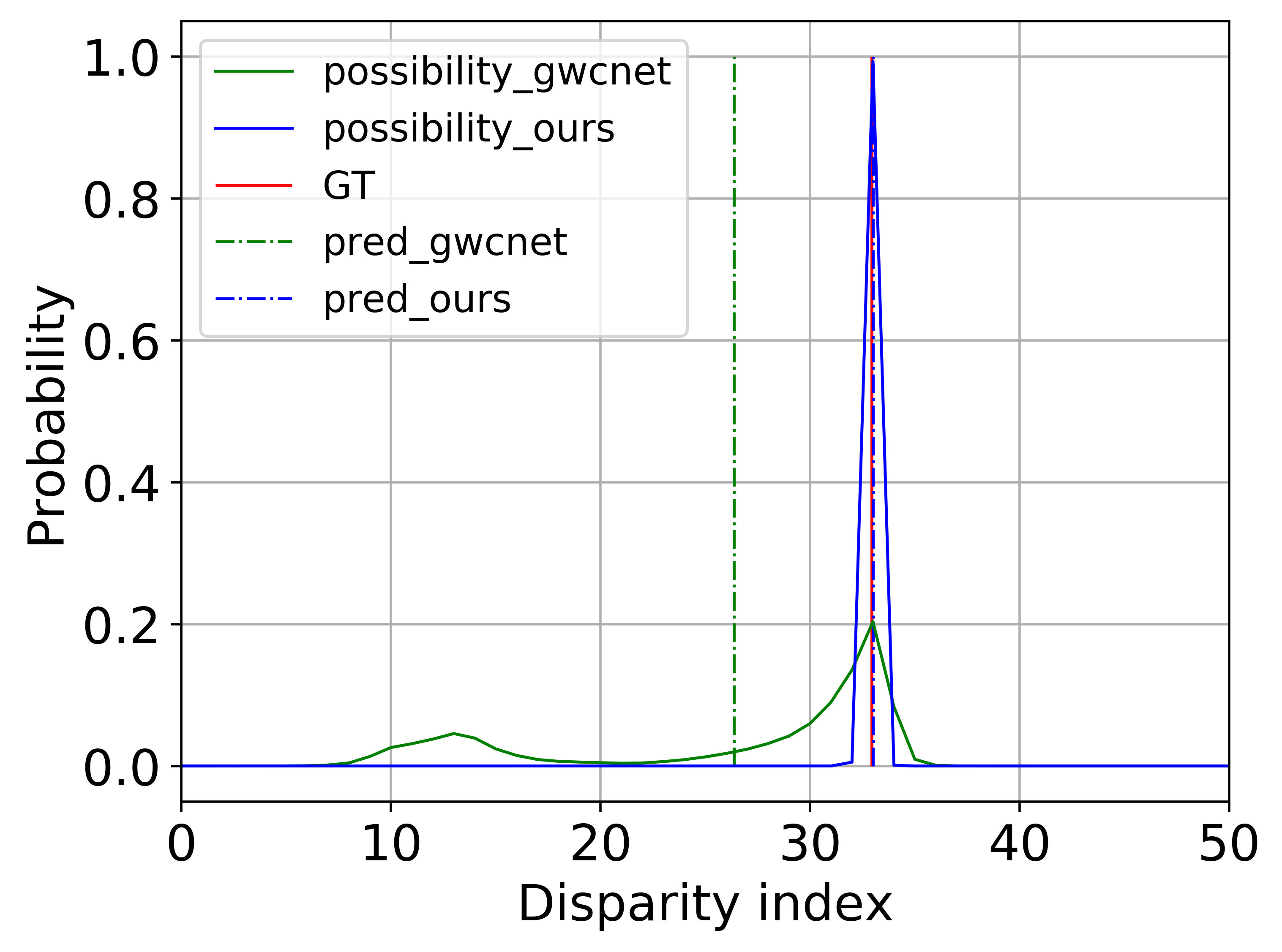}&
     \includegraphics[width=0.4\linewidth]{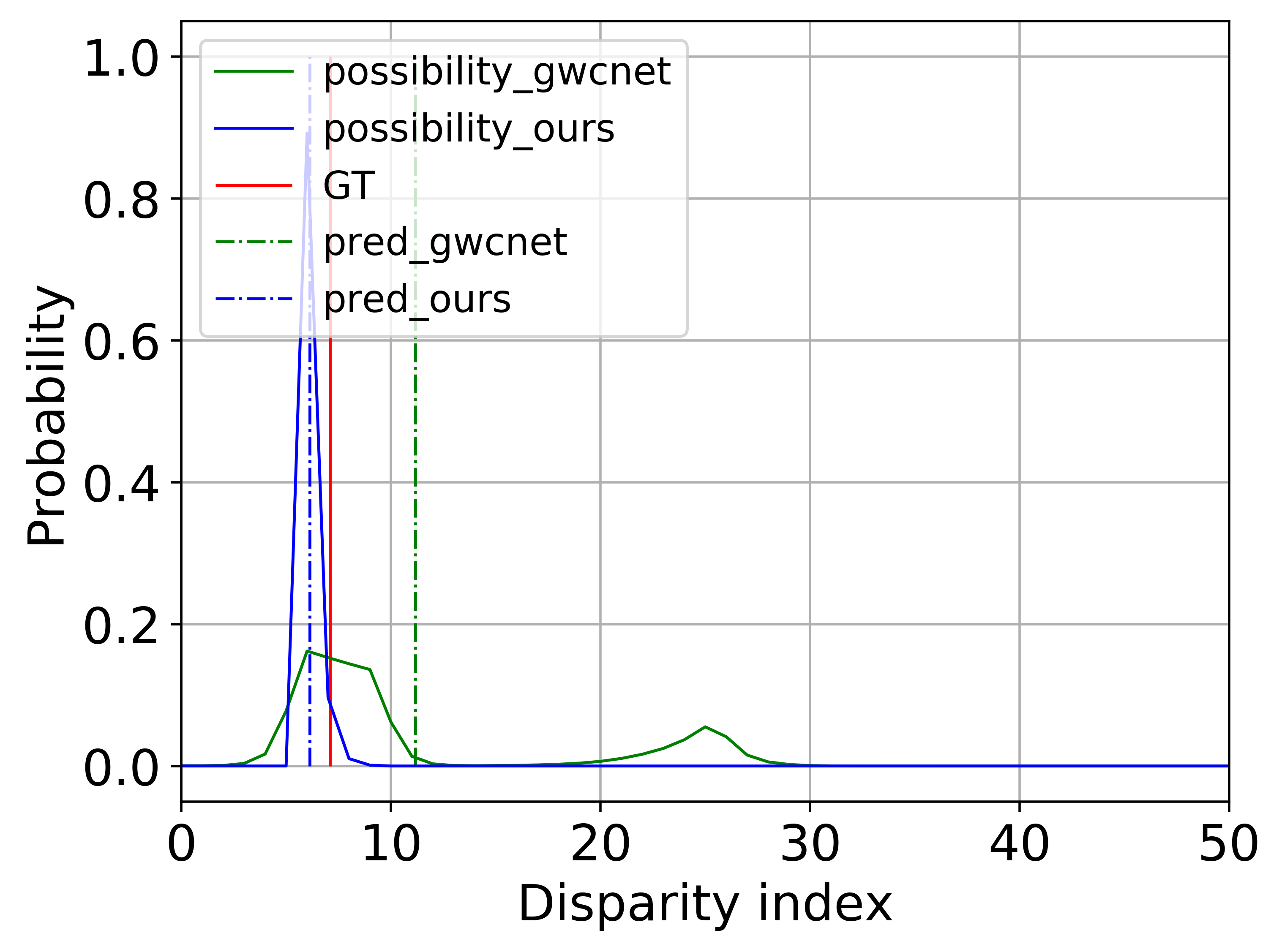}\\
	 \multicolumn{2}{c}{(a)} \\
 	\includegraphics[width=0.4\linewidth]{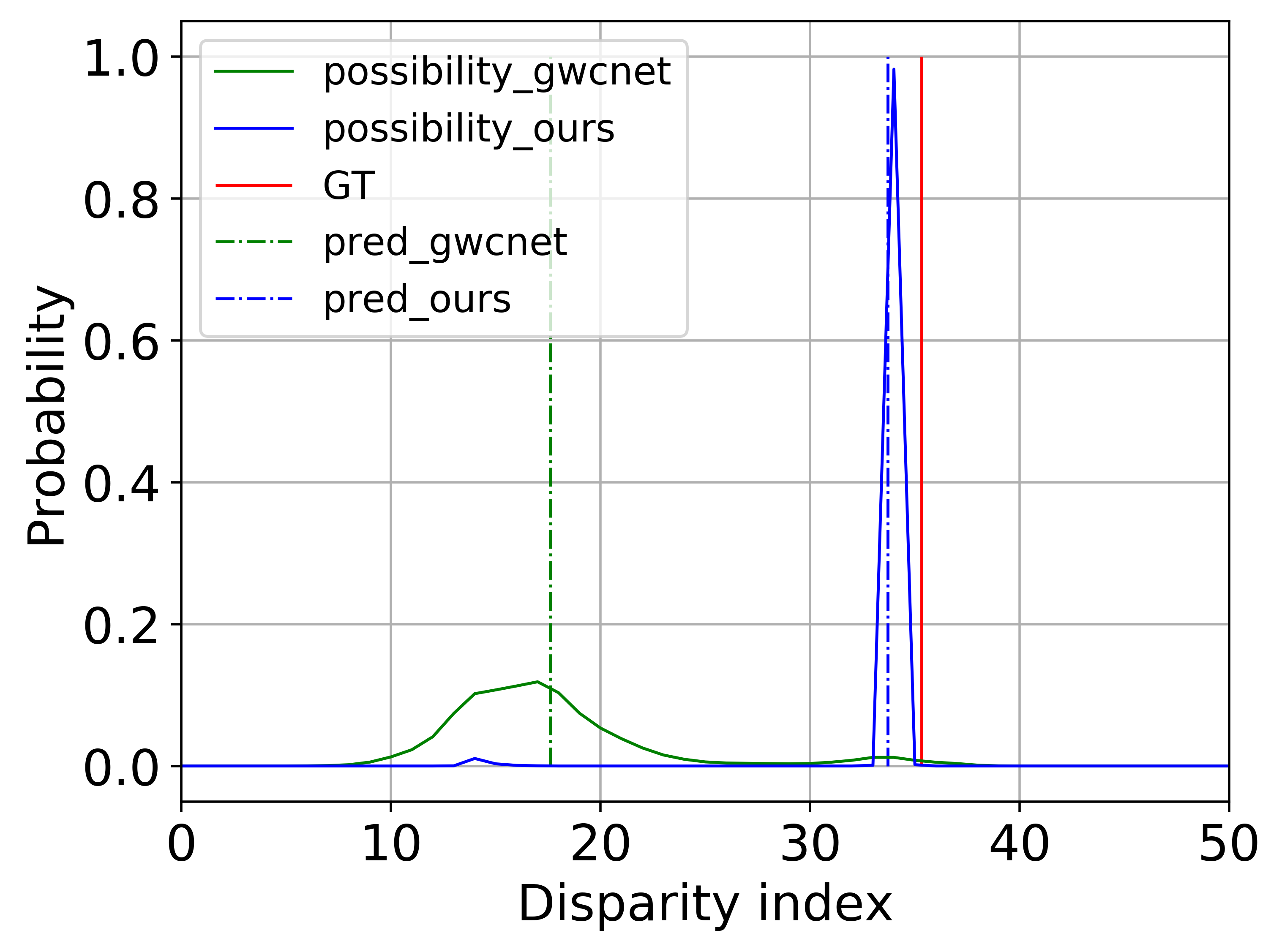}&
     \includegraphics[width=0.4\linewidth]{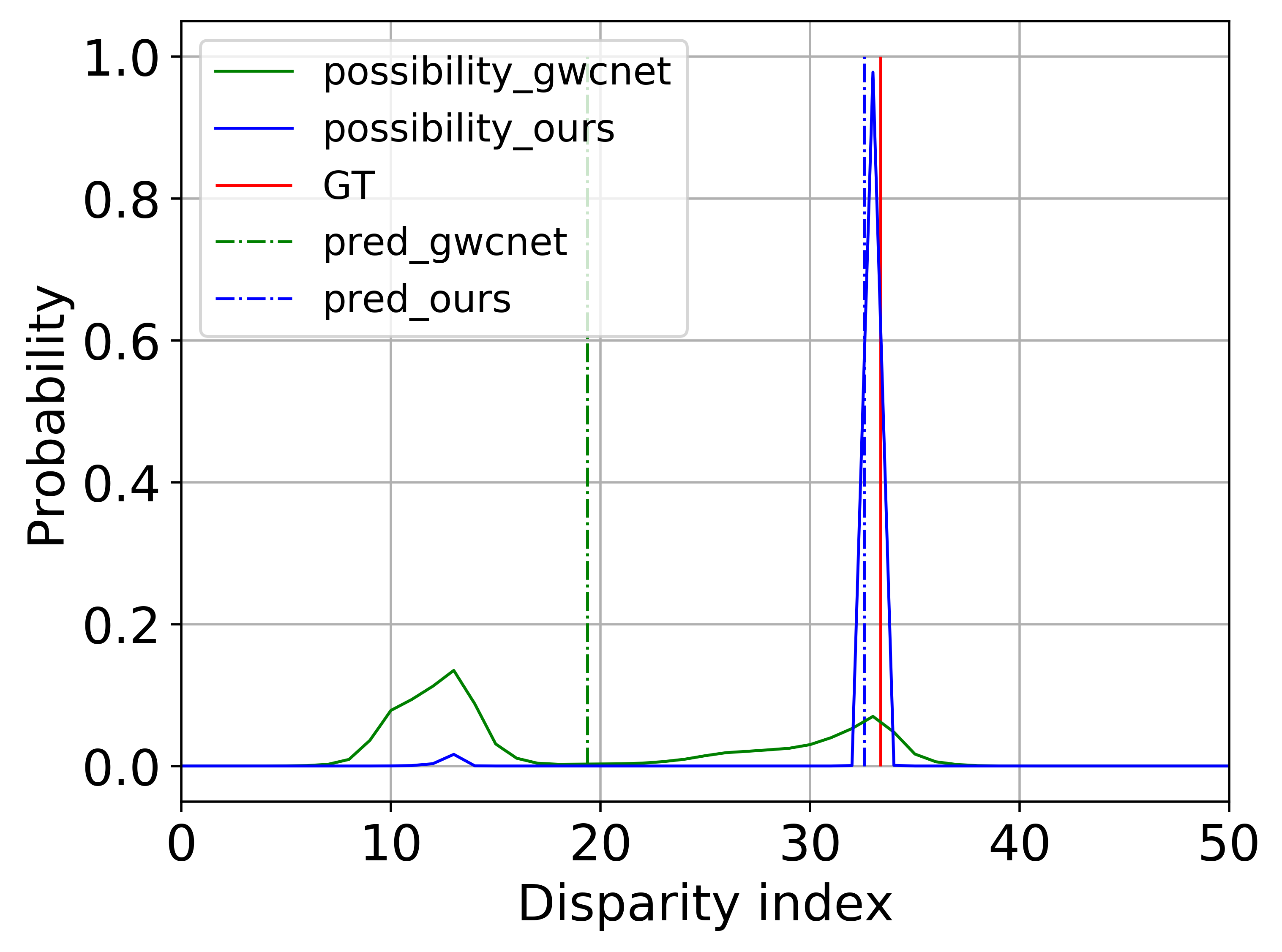}\\
     \multicolumn{2}{c}{(b)} \\
	\end{tabular}
 \vspace{-0.1in}
	\caption{\small More detailed disparity probability distribution comparison between GWCNet and GWCNet\_CMD. All methods are trained on the combination of the SceneFlow dataset and
unlabeled target domain data. Two wrong disparity probability distributions can be corrected by the proposed method: (a) method predicts a multimodal distribution while it can predict a higher probability at the true disparity index. (b) method directly predicts the highest probability at the wrong disparity index while it still can predict some faint probability at the true disparity index.
}
	\label{fig: distribution}
\vspace{-0.2in}
\end{figure}

In this section, we give a more detailed analysis of the disparity probability distribution that can be optimized by the proposed method. All methods are trained on the combination of the SceneFlow dataset and unlabeled target domain data. As shown in Fig. \ref{fig: distribution}, the proposed method can alleviate two wrong disparity probability distributions: (1) Method predicts a multimodal distribution while it can predict a higher probability at the true disparity index (Fig. \ref{fig: distribution}(a)). For this case, we can directly constrain the multi-modal distribution of networks to obtain a more accurate disparity prediction result. (2) Method directly predicts the highest probability at the wrong disparity index while it still can predict some faint probability at the true disparity index (Fig. \ref{fig: distribution}(b)). For this case, we mainly rely on the stronger penalty of temperate softargmin to push the distribution converging to the right distribution, i.e., a unimodal distribution peaked at the ground true disparity (see Sec.3 for more detail). 
Both quantitative (Tab. 1\&3 of the main paper) and qualitative results demonstrate that the proposed method successfully alleviates the issue of multimodal distribution on the target domain and achieves better generalization.

\section{Gradient Analysis Proof}
Following all the variable definitions in the paper, here, we prove Eq. 15 of the main paper:
\begin{eqnarray}
\frac{\partial L_s^{'}}{\partial C(i)} = \frac{\partial L_s^{'}}{\partial p(i)} \frac{\partial p(i)}{\partial C(i)} = t \frac{\partial L_s}{\partial p(i)} \frac{\partial p(i)}{\partial C(i)}
\label{gradient}
\end{eqnarray}
 Let $-C(i) = g$,
\begin{equation}
\begin{aligned}
\frac{\partial L_s^{'}}{\partial g} &= \frac{\partial L_s^{'}}{\partial \hat d_s} \frac{\partial \hat d_s}{\partial p(i)} \frac{\partial p(i)}{\partial f^{'}(g)}\frac{\partial f^{'}(g)}{\partial g}\\
&=\frac{\partial L_s}{\partial p(i)} \frac{\partial p(i)}{\partial f^{'}(g)}\frac{\partial f^{'}(g)}{\partial g}
\end{aligned}
\label{prove}
\end{equation}
Because $f'(x) =e^{tx}$ and $f(x) = e^{x}$, 
\begin{equation}
\frac{\partial f^{'}(x)}{\partial x} \approx  t \frac{\partial f(x)}{\partial x}
\end{equation}
So, 
\begin{eqnarray}
\frac{\partial L_s^{'}}{\partial g} = \frac{\partial L_s^{'}}{\partial p(i)} \frac{\partial p(i)}{\partial g} \approx t \frac{\partial L_s}{\partial p(i)} \frac{\partial p(i)}{\partial g} = t\frac{\partial L_s}{\partial g}
\end{eqnarray}

As shown in Eq.1, compared to $L_s$, $ L_s^{'}$ assigns t times larger gradient to the wrong disparity index. Such stronger penalty can help the method better correct the second wrong distribution in Sec. 2, i.e., directly predicts the highest probability at the wrong disparity index (Fig. \ref{fig: distribution}(b)).
\bibliographystyle{splncs04}
\bibliography{egbib}